\documentclass[a4paper,fleqn]{cas-sc}

\usepackage[numbers]{natbib}

\def\tsc#1{\csdef{#1}{\textsc{\lowercase{#1}}\xspace}}
\tsc{WGM}
\tsc{QE}
\tsc{EP}
\tsc{PMS}
\tsc{BEC}
\tsc{DE}

\renewcommand{\arraystretch}{1.15}

\usepackage{tabularx}
\usepackage{array}
\usepackage{booktabs}

\begin{document}

\makeatletter
\def\ps@pprintTitle{%
  \let\@oddhead\@empty
  \let\@evenhead\@empty
  \let\@oddfoot\@empty
  \let\@evenfoot\@empty
}
\makeatother

\let\WriteBookmarks\relax
\def\floatpagepagefraction{1}
\def\textpagefraction{.001}

\shorttitle{Smart Classroom Behavior Analysis with HCCB}
\shortauthors{Xu et~al.}

\title[mode = title]{A Smart Classroom Behavior Analysis Framework with a New Highly Congested Classroom Dataset}
\tnotemark[1]

\tnotetext[1]{This work was supported by the Natural Science Foundation of Liaoning Province (Nos. 2025-MS-134 and 2022-MS-353) and the Basic Scientific Research Project of the Education Department of Liaoning Province (Nos. LJ232410146053 and LJKMZ20220640).}
\author[1]{Wei Xu} \ead{24210811000473@stu.ustl.edu.cn} \credit{Conceptualization, Methodology, Software, Data curation, Formal analysis, Validation, Visualization, Writing -- original draft}
\author[1]{Maoxiang Chu} \cormark[1] \ead{chu52_2004@163.com} \credit{Supervision, Conceptualization, Methodology, Project administration, Resources, Writing -- review and editing} 
\author[1]{Yuelong Fan} \ead{fyl010912@163.com} \credit{Methodology, Software, Data curation, Validation, Formal analysis, Writing -- review and editing} 
\author[1]{Guanghao Liao} \ead{24210811000450@stu.ustl.edu.cn} \credit{Software, Data curation, Validation, Visualization, Writing -- review and editing} 
\author[1]{Yinxiang Yu} \ead{xiangzi_pull_car@163.com} \credit{Data curation, Software, Validation, Visualization, Writing -- review and editing} 
\author[1]{Zhi Chen} \ead{cz1073755937@gmail.com} \credit{Software, Data curation, Validation, Formal analysis, Writing -- review and editing} 
\author[1]{Haotian Wang} \ead{wanyanwang59@gmail.com} \credit{Data curation, Software, Validation, Visualization, Writing -- review and editing} 
\author[1]{Yutian Zhu} \ead{ytz1024201@163.com} \credit{Data curation, Software, Validation, Visualization, Writing -- review and editing} 
\affiliation[1]{ organization={University of Science and Technology Liaoning}, addressline={School of Electronic and Information Engineering}, postcode={114051}, city={Anshan}, country={China}}
\cortext[cor1]{Corresponding author.}

\begin{abstract}
Student behavior detection plays an important role in intelligent classroom analysis, but remains challenging in real-world large-class scenarios due to dense instance co-occurrence, asymmetric occlusion, depth-wise scale discontinuity, and fine-grained semantic degradation in distant targets. Existing classroom behavior datasets and general-purpose detectors are insufficient to characterize and address these structural challenges. To this end, this paper constructs the Highly Congested Classroom Behavior (HCCB) dataset, covering 50,229 student behavior instances across seven behavior categories: reading, writing, heads up, sleeping, looking around, bowing head, and using phone. HCCB provides a challenging benchmark by integrating dense distributions, severe occlusion, depth-wise scale variations, and fine-grained behavioral semantics into a unified detection setting. To address these challenges, this paper proposes ODER-HSFNet, a YOLO-based detection framework tailored to highly crowded classrooms. At the core of ODER-HSFNet, we design three task-specific innovations to address the structural difficulties of highly crowded classroom behavior detection. First, the Occlusion-aware Deformable Edge Rectifier (ODER) enhances boundary evidence through bounded edge resampling and topology-aware routing, alleviating occlusion and visible-information loss in distant regions. Second, the Hypergraph-State Spatial Fusion (HSSF) module combines local structure enhancement, state-space contextual modeling, and high-order hypergraph relation aggregation to reduce dense instance competition and cross-scale feature misalignment. Finally, the Occlusion-Calibrated Detection Head (OCDetect) suppresses low-quality Pre-NMS candidate boxes through objectness calibration, reducing false positives caused by occlusion boundaries and neighboring instances. The experimental results on two classroom behavior detection datasets demonstrate that the proposed ODER-HSFNet outperforms mainstream YOLO-series methods in terms of \(\mathrm{mAP}_{50:95}\) and \(\mathrm{mAP}_{50}\) metrics, achieving 60.60\%/80.12\% on HCCB and 57.36\%/74.65\% on SCB-D3-S. Further ablation experiments demonstrate the effectiveness of the proposed method in highly crowded classroom behavior detection.
\end{abstract}

\begin{graphicalabstract}
\IfFileExists{figures/graphical_abstract.pdf}{%
  \includegraphics[width=.8\linewidth]{figures/graphical_abstract.pdf}%
}{}
\end{graphicalabstract}
\begin{highlights} 
\item A new highly congested classroom behavior dataset is constructed to benchmark dense, occluded, and depth-varying smart classroom scenarios. 
\item ODER-HSFNet improves classroom behavior detection through occlusion-aware boundary rectification, high-order cross-scale fusion, and confidence calibration. 
\item Extensive experiments on HCCB and SCB-D3-S demonstrate superior accuracy and robustness over mainstream YOLO-series detectors. 
\end{highlights} 
\maketitle

\section{Introduction}\label{sec:introduction}
With the rapid development of digital education technologies and artificial intelligence, smart classrooms have become an important component of modern education systems~\cite{zhou2022stuart}. By automatically detecting and analyzing student behaviors from classroom videos, smart classroom systems can provide teachers with feedback on classroom engagement, learning states, and abnormal behaviors, thereby supporting teaching evaluation, classroom management, and personalized educational decision-making~\citep{lin2021student}. Accordingly, student behavior detection in classroom scenarios has emerged as an important research topic at the intersection of educational intelligence and computer vision~\citep{zhou2022stuart,lin2021student,yang2023scbdataset3,yang2023studentyolov7}.

Different from generic object detection, classroom behavior detection requires not only localizing each student instance but also recognizing its behavior category, such as \texttt{heads up}\texttt{bowing head}\texttt{looking around}\texttt{using phone}~\citep{yang2023scbdataset3,yang2023studentyolov7}. However, in real large-scale classrooms, students are typically distributed densely across multiple rows. This setting introduces severe inter-instance occlusion, significant scale differences between front and rear rows, and substantial degradation of local behavioral cues for distant students due to low resolution, blur, and occlusion. As a result, real classroom behavior detection faces four major challenges.

First, extremely high instance density leads to persistent detection competition. A single image from a large-scale classroom often contains dozens or even hundreds of student instances. Dense targets may interfere with local foreground responses, resulting in merged adjacent bounding boxes, missed detections of small rear-row targets, and unstable positive sample assignment. Second, asymmetric front-to-rear occlusion weakens the visible evidence of rear-row targets. Classroom occlusion is usually not caused by random overlap, but by directional coverage induced by seat arrangement, camera viewpoint, and large foreground targets. Consequently, rear-row small targets often retain only partial cues, such as the head, shoulders, hands, or associated objects. Third, the depth structure of multi-row lecture classrooms causes cross-scale feature mismatch. Front-row targets are relatively large, whereas rear-row targets are much smaller. Although conventional feature pyramids can handle general scale variations~\citep{lin2017fpn}, they remain insufficient for simultaneously preserving the semantic stability of large foreground targets and the detail fidelity of small rear-row targets under continuous front-to-rear scale discontinuities. Fourth, far-field degradation compresses the semantic boundaries among fine-grained behavior categories. Behaviors such as \texttt{reading}, \texttt{writing}, \texttt{bowing head}, and \texttt{using phone} often depend on local evidence, including paper regions, pen tips, hand movements, or visible object areas. Such cues are easily lost for distant small targets because of occlusion and low resolution, which further aggravates category confusion.

Early classroom behavior analysis methods mainly relied on handcrafted features, pose estimation, or traditional machine learning models~\citep{lin2021student}. These methods can be effective in low-density and controlled scenarios, but their generalization ability is limited when dealing with dense targets, severe occlusion, and drastic scale variations in real large-scale classrooms. In recent years, with the development of deep learning and object detection techniques, methods such as the YOLO series~\citep{redmon2016yolo,lei2025yolov13}, Faster R-CNN~\citep{ren2015faster}, SSD~\citep{liu2016ssd}, and Transformer-based detectors~\citep{carion2020detr} have been introduced into classroom behavior detection~\citep{yang2023scbdataset3}, achieving a favorable balance between accuracy and efficiency. Nevertheless, most existing methods are evaluated on conventional classroom datasets~\citep{yang2023studentyolov7}, which mainly emphasize behavior category definitions while providing insufficient coverage of extremely high density, asymmetric occlusion, and depth-induced scale conflicts. Therefore, stable performance on conventional classroom benchmarks does not necessarily indicate that a detector can handle the complex spatial geometry of real large-scale classrooms.

To address the insufficient modeling of highly congested scenarios in existing classroom behavior detection datasets, this paper constructs the Highly-Congested Classroom Behavior Dataset, abbreviated as HCCB. The dataset is collected from real multi-row lecture classrooms and contains 796 high-resolution images with 50,229 student behavior instances, covering seven behavior categories: \texttt{reading}, \texttt{writing}, \texttt{heads up}, \texttt{sleeping}, \texttt{looking around}, \texttt{bowing head}, and \texttt{using phone}. Compared with existing classroom behavior datasets, the main value of HCCB lies not merely in increasing the number of samples, but in integrating extremely high instance density, severe object-level occlusion, significant front-to-rear scale discontinuities, and fine-grained classroom behavior semantics into a unified detection benchmark. It therefore provides a more challenging evaluation platform for student behavior detection in real large-scale classrooms.

To address the structural challenges revealed by HCCB, this paper further proposes ODER-HSFNet, a YOLO-based framework for highly congested classroom behavior detection. The framework consists of three complementary components, namely ODER, HSSF, and OCDetect, which enhance detector robustness from three perspectives: local visible evidence compensation, cross-scale high-order relation fusion, and detection-side candidate-box calibration. Specifically, the Occlusion-aware Deformable Edge Rectifier (ODER) performs controlled compensation of reliable visible evidence around occlusion boundaries and far-field degraded regions through bounded deformable edge resampling~\citep{dai2017deformable}, topology-aware sampling routing, and sample-level residual amplitude modulation. The Hypergraph-State Spatial Fusion (HSSF) module alleviates cross-layer feature mismatch caused by dense instance competition and depth-induced scale discontinuities through local structure enhancement, state-space context modeling~\citep{gu2023mamba}, and high-order hypergraph relation fusion~\citep{feng2019hypergraph}. The Occlusion-Calibrated Detection Head (OCDetect) introduces a class-agnostic objectness calibration branch at the detection stage to suppress low-quality candidate boxes generated by desk and chair edges, partial limbs, and occlusion boundaries before NMS.

We conduct systematic experiments on HCCB and the public SCB-Dataset3-S (SCB-D3-S) dataset~\citep{yang2023scbdataset3}. HCCB is used as the primary benchmark to evaluate the structural robustness of the proposed model in real highly congested large-scale classrooms, while SCB-D3-S is used as an external public dataset to analyze its generalization ability in conventional classroom behavior detection scenarios.

The main contributions are summarized as follows.

\begin{enumerate}
\item A new HCCB dataset is constructed for highly congested classroom behavior detection. Collected from real large-scale lecture classrooms, HCCB contains 796 high-resolution images and 50,229 student behavior instances, covering seven classroom behaviors: \texttt{reading}, \texttt{writing}, \texttt{heads up}, \texttt{sleeping}, \texttt{looking around}, \texttt{bowing head}, and \texttt{using phone}. Compared with existing classroom behavior datasets, HCCB reveals the challenges of high instance density, severe object-level occlusion, significant depth-induced scale discontinuities, and fine-grained behavior semantic confusion in real classroom student behavior detection, while providing a systematic and challenging benchmark for highly congested classroom behavior detection.

\item An ODER module is proposed for occlusion-edge evidence compensation. Through bounded deformable edge sampling, topology-aware sampling routing, and sample-level residual amplitude modulation, ODER adaptively re-aggregates reliable visible evidence in asymmetric occlusion and far-field degraded regions. This design alleviates feature contamination around occlusion boundaries, interference from neighboring instances, and missed detections of small rear-row targets.

\item An HSSF module is designed for cross-scale high-order feature fusion. HSSF integrates DSConv-based local structure enhancement, VSS-Mamba-based long-range context modeling, and adaptive hypergraph-based high-order relation aggregation to mitigate cross-layer feature mismatch caused by dense instance competition and depth-induced scale discontinuities.

\item An OCDetect head is proposed for occlusion-calibrated detection. In addition to the classification and regression branches, OCDetect introduces class-agnostic objectness calibration to suppress low-quality candidate boxes generated by desk and chair edges, partial limbs, and occlusion transition regions before NMS. This improves the ranking quality of candidate boxes in highly congested classroom scenarios at the detection output level.

\end{enumerate}

Based on these contributions, systematic experiments are conducted on HCCB and the public SCB-D3-S dataset, including comparisons with mainstream detectors, validation of dataset benchmark difficulty, candidate-box quality analysis, occluded-target evaluation, distant small-target evaluation, and module ablation studies. The experimental results demonstrate the effectiveness of ODER-HSFNet for highly congested classroom behavior detection, and further analyze its generalization performance and applicability boundaries in conventional classroom behavior detection scenarios.

\section{Related Work}\label{sec:related_work}

\subsection{Student Behavior Detection in Smart Classrooms}\label{subsec:student_behavior_detection}
Student behavior detection is a fundamental task in smart classroom perception, teaching-state assessment, and classroom feedback analysis~\citep{yang2023studentyolov7}. Its objective is not only to determine whether a certain type of behavior is present in the classroom, but also to localize specific student instances in images or videos and recognize their corresponding behavior categories~\citep{yang2023scbdataset3}. Therefore, unlike general image classification or holistic classroom-state recognition, student behavior detection emphasizes instance-level localization, fine-grained semantic discrimination, and robustness in complex classroom environments.

Early classroom behavior analysis methods mainly relied on handcrafted features, pose estimation, or traditional machine learning models. These methods inferred student states from head orientation, human posture, hand movements, and local appearance cues~\citep{lin2021student}. They can be effective in low-density and controlled classroom environments, but are sensitive to occlusion, scale variation, similar postures, and far-field blur, which limits their applicability to real large-scale classrooms. With the development of deep learning, convolutional neural networks, object detection networks, pose modeling methods, and Transformer-based methods have been gradually introduced into classroom behavior analysis~\citep{liu2016ssd,redmon2016yolo,carion2020detr}, leading to clear improvements in recognition accuracy and real-time performance.

In recent years, object-detection-based classroom behavior analysis has received increasing attention. Frameworks such as Faster R-CNN~\citep{ren2015faster}, SSD~\citep{liu2016ssd}, the YOLO series~\citep{redmon2016yolo,lei2025yolov13}, and Transformer-based detectors~\citep{carion2020detr} can simultaneously perform student instance localization and behavior classification. They are therefore more consistent with the requirement for individual-level state perception in smart classrooms. Existing studies have commonly improved classroom behavior detection performance through attention mechanisms, multi-scale feature fusion, lightweight convolution, or detection-head optimization~\citep{yang2023scbdataset3,yang2023yolov7bra}. Among these methods, the YOLO series is widely used in real-time or near-real-time classroom behavior detection because of its favorable balance between speed and accuracy~\citep{yang2023studentyolov7}.

However, most existing classroom behavior detection studies focus on conventional classrooms or medium- and low-density scenarios. Their main concerns are usually behavior category recognition, small object detection, or lightweight model design~\citep{yang2023yolov7bra}. Systematic modeling is still insufficient for the concurrent challenges in real large-scale classrooms, including high instance density, asymmetric front-to-rear occlusion, depth-induced scale discontinuities, and far-field fine-grained semantic degradation. In multi-row lecture classrooms, rear-row students are often occluded by front-row targets and desk-chair structures. Meanwhile, behaviors such as reading, writing, bowing head, and using phone rely on local cues, including hands, paper regions, head posture, and associated objects. As a result, conventional detectors are prone to missed detections, false detections, and category confusion. Therefore, classroom behavior detection requires not only more challenging data benchmarks, but also specialized model designs for structural failure modes in highly congested classroom scenarios.

\subsection{Object Detection under Dense Occlusion}\label{subsec:dense_occlusion_detection}
Object detection in dense occlusion scenarios is an important problem in computer vision and is widely encountered in complex visual applications, such as pedestrian detection, crowd analysis, traffic surveillance, and remote sensing~\citep{shao2018crowdhuman,zhang2017citypersons,zhang2020widerperson,sindagi2020jhucrowd}. Compared with conventional object detection, targets in dense occlusion scenarios usually exhibit small inter-object distances, high bounding-box overlap, incomplete visible regions, and imbalanced scale distributions. These factors jointly affect classification discrimination, bounding-box regression, positive sample assignment, and post-processing, making detectors prone to adjacent instance merging, missed detections of small objects, candidate-box congestion, and erroneous suppression by NMS~\citep{bodla2017softnms,liu2019adaptive,wang2018repulsion,chu2020crowded}.

To alleviate the challenges of dense detection, existing studies have mainly focused on multi-scale feature fusion, occlusion-robust localization, sample assignment optimization, candidate quality estimation, and post-processing improvement. FPN~\citep{lin2017fpn}, PANet~\citep{liu2018panet}, and BiFPN~\citep{tan2020efficientdet} enhance the representation of objects at different scales through cross-layer feature propagation, and have become widely used multi-scale modeling strategies for dense detection. Cascade R-CNN~\citep{cai2018cascade} improves candidate-box quality by progressively increasing IoU thresholds. ATSS~\citep{zhang2020atss} and OTA~\citep{ge2021ota} optimize positive-negative sample matching from the perspectives of adaptive sample selection and global optimal assignment, respectively. GFL~\citep{li2020gfl} combines classification confidence with localization quality modeling to improve the ranking reliability of dense candidate boxes. For occlusion and deformation, deformable convolution~\citep{dai2017deformable} and deformable attention~\citep{zhu2021deformabledetr} enable models to dynamically adjust their receptive fields according to object shapes and local structures by learning spatial sampling offsets. Repulsion Loss~\citep{wang2018repulsion} reduces localization shifts in crowded pedestrian detection by imposing repulsive constraints between neighboring objects, while CrowdDet~\citep{chu2020crowded} improves the recall of highly overlapping instances through a one-proposal multiple-predictions mechanism. In addition, Soft-NMS~\citep{bodla2017softnms} and Adaptive NMS~\citep{liu2019adaptive} mitigate erroneous suppression of highly overlapping objects at the post-processing stage.

Although these methods improve detection performance in dense scenarios, they cannot fully address the structural challenges in highly congested classrooms. First, classroom occlusion is not random overlap, but directional occlusion jointly induced by seat arrangement, camera viewpoint, and the spatial relationship between front and rear rows. Large foreground targets in the front rows often partially occlude small targets in the rear rows, leaving rear-row instances with only limited visible evidence, such as the head, shoulders, hands, or associated objects. Second, unconstrained dynamic sampling may cross into neighboring instances or desk-chair background regions in dense student areas, thereby introducing feature contamination. Finally, NMS and its variants~\citep{bodla2017softnms,liu2019adaptive} mainly adjust detection scores according to candidate-box overlap, making it difficult to actively distinguish real student instances from low-quality candidate boxes generated by occlusion boundaries, partial limbs, and background textures.

Therefore, highly congested classroom behavior detection needs to go beyond the processing paradigm of generic dense object detection. The model should not only recognize dense targets, but also re-aggregate reliable visible evidence under asymmetric occlusion and suppress false candidate boxes induced by desk-chair edges, partial limbs, and occlusion transition regions at the detection stage.

\subsection{Cross-scale and High-order Context Modeling}\label{subsec:cross_scale_high_order_context}
Cross-scale feature fusion is a fundamental issue in object detection. Shallow features usually preserve richer spatial details and are beneficial for small object localization, whereas deep features provide stronger semantic representations for large objects and complex category recognition. FPN~\citep{lin2017fpn} and its variants, such as PANet~\citep{liu2018panet} and BiFPN~\citep{tan2020efficientdet}, alleviate the influence of scale variation on detection performance by fusing features from different levels. PVT~\citep{wang2021pvt} and Swin Transformer~\citep{liu2021swin} further introduce hierarchical structures into Transformer backbones, making them more suitable for the multi-scale representation requirements of dense prediction tasks. However, in real multi-row lecture classrooms, scale variation is not a random perturbation, but a front-to-rear scale discontinuity continuously induced by the classroom depth structure. Front-row targets are relatively large, whereas rear-row targets are smaller and often suffer from occlusion and low-resolution degradation. In this situation, conventional progressive fusion may smooth out the edge details of small targets and introduce semantic misalignment between shallow and deep features.

To enhance long-range dependency modeling, Transformer-based detectors~\citep{carion2020detr,zhu2021deformabledetr}, Non-local modules~\citep{wang2018nonlocal}, and visual state-space models~\citep{gu2023mamba,liu2024vmamba} have been introduced into visual detection and dense prediction tasks. Transformers~\citep{carion2020detr} capture global contextual relationships through self-attention, but their computational cost is high in high-resolution dense detection scenarios. Deformable DETR~\citep{zhu2021deformabledetr} reduces the computational burden of global attention through sparse deformable attention, while PVT~\citep{wang2021pvt} and Swin Transformer~\citep{liu2021swin} improve the applicability of visual Transformers to dense tasks through hierarchical structures and local window mechanisms. In contrast, state-space modeling methods such as Mamba~\citep{gu2023mamba} and VMamba~\citep{liu2024vmamba} provide a new perspective for long-sequence visual feature modeling, enabling broad contextual interaction with lower complexity. For highly congested classroom images, long-range context helps distinguish student instances with similar local appearances and compensates for evidence loss caused by local occlusion.

In addition to long-range dependencies, high-order relation modeling is also important for dense classroom detection. Relation Networks~\citep{hu2018relation} demonstrate that inter-object relation modeling can enhance instance interaction during detection. Graph neural networks~\citep{kipf2017gcn} model dependencies between objects or regions through node connections, while hypergraph neural networks~\citep{feng2019hypergraph} further extend pairwise relations to multi-node high-order relations. This makes it possible to describe joint semantics among regions with similar depth, similar behavior patterns, or local student groups. In classroom scenarios, student instances are usually distributed in multi-row layouts, where strong spatial and occlusion correlations exist among neighboring targets. Relying only on local convolution or pairwise relation modeling is insufficient to fully characterize such group structures.

Therefore, highly congested classroom behavior detection needs to jointly consider local detail preservation, long-range context modeling, and high-order relation aggregation. Local structures help recover fine-grained cues such as hands, heads, and associated objects. State-space modeling enhances cross-region contextual perception, while high-order relation aggregation alleviates response competition among dense students and cross-scale semantic mismatch. Existing methods rarely model these three types of information in a unified manner for real highly congested classroom behavior detection, leaving room for further model design.

\subsection{Classroom Behavior and Dense Crowd Datasets}\label{subsec:classroom_behavior_dense_datasets}
Whether a dataset can sufficiently cover task-specific challenges is a key factor in driving the development of detection algorithms. Existing related datasets can be broadly categorized into two groups: classroom behavior datasets and generic dense crowd datasets. Classroom behavior datasets, such as StuArt \citep{zhou2022stuart}, SCB-Dataset3 \citep{yang2023scbdataset3}, and the SCB series classroom behavior datasets \citep{yang2023studentyolov7,yang2023yolov7bra}, usually provide explicit educational semantic labels, including reading, writing, raising hands, bowing head, sleeping, and using phone. These datasets support research on classroom state perception and student behavior recognition. In contrast, generic dense crowd datasets, such as CityPersons \citep{zhang2017citypersons}, CrowdHuman \citep{shao2018crowdhuman}, WiderPerson \citep{zhang2020widerperson}, and JHU-CROWD++ \citep{sindagi2020jhucrowd}, focus more on high-density human instance detection, crowded pedestrian separation, or dense head annotation, and are suitable for evaluating localization and separation abilities in crowded open scenarios.

Existing classroom behavior datasets have advanced smart classroom behavior analysis. However, most of them are collected in relatively conventional teaching environments, where target density, occlusion severity, and depth-induced scale conflicts are limited \citep{zhou2022stuart,yang2023scbdataset3,yang2023studentyolov7,yang2023yolov7bra}. These datasets are suitable for evaluating basic classroom behavior detectability, but are insufficient for assessing the structural robustness of detectors in real large-scale classrooms. In multi-row lecture classrooms, student instances appear densely and concurrently. The scale difference between front and rear rows is significant, rear-row small targets are persistently occluded by front-row targets and desk-chair structures, and fine-grained behavior categories strongly depend on local visual evidence. These challenges are usually not systematically represented in existing classroom behavior datasets.

Generic dense crowd datasets are more challenging in terms of target crowding, occlusion, and instance separation \citep{shao2018crowdhuman,zhang2017citypersons,zhang2020widerperson,sindagi2020jhucrowd}. Nevertheless, their target semantics are relatively limited, as they mainly focus on human body, pedestrian, or head detection and do not involve fine-grained semantic discrimination of classroom behaviors. Moreover, the viewpoints, spatial layouts, and occlusion patterns of generic crowd datasets differ substantially from those of real classrooms. As a result, they cannot adequately reflect the front-to-rear depth structure, behavior category confusion, and educational semantic requirements in classroom scenarios. Therefore, directly applying generic dense crowd datasets to classroom behavior detection cannot comprehensively evaluate model applicability in real smart classrooms.

These observations indicate a clear mismatch between dataset properties and task requirements. Classroom behavior datasets provide educational semantics but lack extreme crowding and strong occlusion structures, whereas generic dense crowd datasets provide spatial crowding but lack classroom behavior semantics and depth-related scale regularities. Because of this mismatch, a model that performs well on existing datasets may still suffer from missed detections of rear-row small targets, failed separation of occluded targets, increased local candidate-box noise, and fine-grained behavior category confusion in real large-scale classrooms.

Therefore, for real smart classroom applications, it is necessary to construct a dataset that simultaneously contains high instance density, severe asymmetric occlusion, significant depth-induced scale discontinuities, and fine-grained classroom behavior semantics. Such a dataset can serve not only as a stress-test benchmark for highly congested classroom behavior detection, but also as a clearer empirical basis for designing methods related to occlusion evidence compensation, cross-scale high-order relation modeling, and detection-side confidence calibration. Based on this consideration, this paper constructs the Highly-Congested Classroom Behavior Dataset, abbreviated as HCCB, and further analyzes its data structure characteristics and detection bottlenecks in the following sections.

\section{The Highly Congested Classroom Behavior Dataset}\label{sec:hccb_dataset}
Based on the analysis in Section 2, existing classroom behavior datasets \citep{zhou2022stuart,yang2023scbdataset3} provide educationally meaningful behavior labels \citep{yang2023studentyolov7,yang2023yolov7bra}, but their coverage of high instance density, asymmetric occlusion, depth-induced scale discontinuities, and far-field fine-grained semantic degradation in real large-scale classrooms remains insufficient. To address this gap, this paper constructs the Highly-Congested Classroom Behavior Dataset, abbreviated as HCCB. The dataset is collected from real multi-row lecture classrooms and is designed for student behavior detection in highly congested classroom scenarios. It aims to integrate high-density instance distributions \citep{shao2018crowdhuman}, severe occlusion \citep{zhang2017citypersons,zhang2020widerperson}, significant front-to-rear scale differences, and fine-grained classroom behavior semantics \citep{yang2023scbdataset3,yang2023yolov7bra} into a unified detection benchmark.
\begin{figure}
\centering
\includegraphics[width=.92\linewidth]{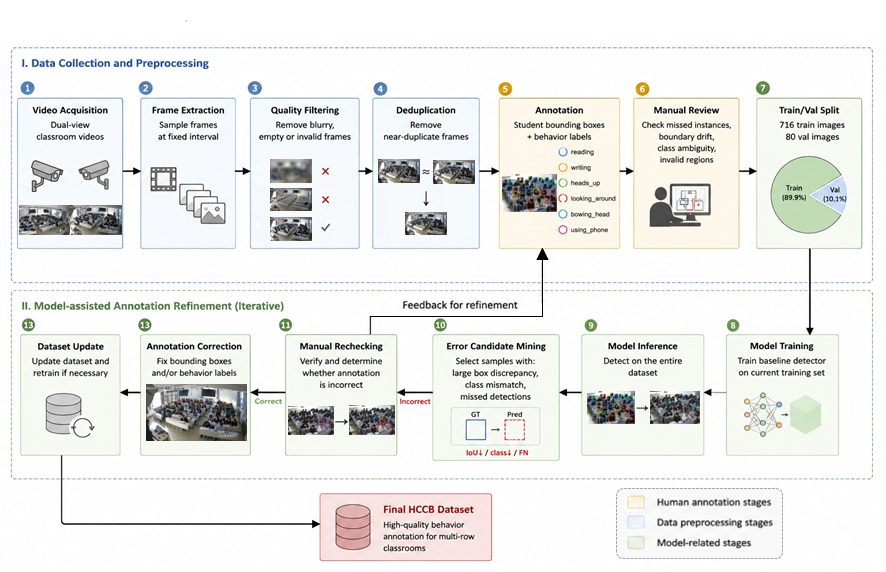}
\caption{HCCB dataset construction and model-assisted annotation refinement pipeline. The yellow modules represent the manual annotation stage, the blue modules represent the data preprocessing stage, and the green modules represent the model-assisted annotation optimization stage.}
\label{fig:hccb_pipeline}
\end{figure}
The construction pipeline of HCCB is illustrated in Fig.~1. The dataset construction process consists of two stages. The first stage is data collection and preprocessing, including dual-view classroom video acquisition, fixed-interval frame extraction, low-quality frame filtering, near-duplicate frame removal, manual annotation, manual review, and training/validation split. The second stage is model-assisted annotation refinement. Through initial detector training, full-dataset inference, erroneous candidate mining, manual rechecking, and annotation correction, missed annotations, category ambiguities, and bounding-box deviations are iteratively corrected. This pipeline improves the annotation consistency for small and occluded targets in highly dense scenarios while maintaining the reliability of manual annotations.
\subsection{Data Acquisition}\label{subsec:data_acquisition}
The raw data of HCCB were collected from a real multi-row tiered lecture classroom at a university. This scenario exhibits a typical large-scale classroom structure: students are densely distributed across multiple rows, clear perspective-induced scale differences exist between front and rear rows, and rear-row targets are often jointly occluded by front-row students, desk-chair structures, and viewpoint constraints. Therefore, this scenario provides a representative setting for studying spatial crowding and depth-related degradation in student behavior detection for real smart classrooms.

To obtain more complete spatial coverage of the classroom, two Hikvision DS-2CD3T45-I3 high-definition bullet cameras were deployed on the left and right sides at the front of the classroom, forming a dual-view acquisition setup, as shown in Fig.~2. Compared with single-view acquisition, the dual-view setup covers more front- and rear-row regions and partially alleviates information loss caused by seat arrangement and human occlusion under a single viewpoint. Both views are oriented toward the student seating area, allowing the dataset to preserve large front-row targets, small middle- and rear-row targets, and occlusion structures across different depth regions.

\begin{figure}
\centering
\includegraphics[width=.98\linewidth]{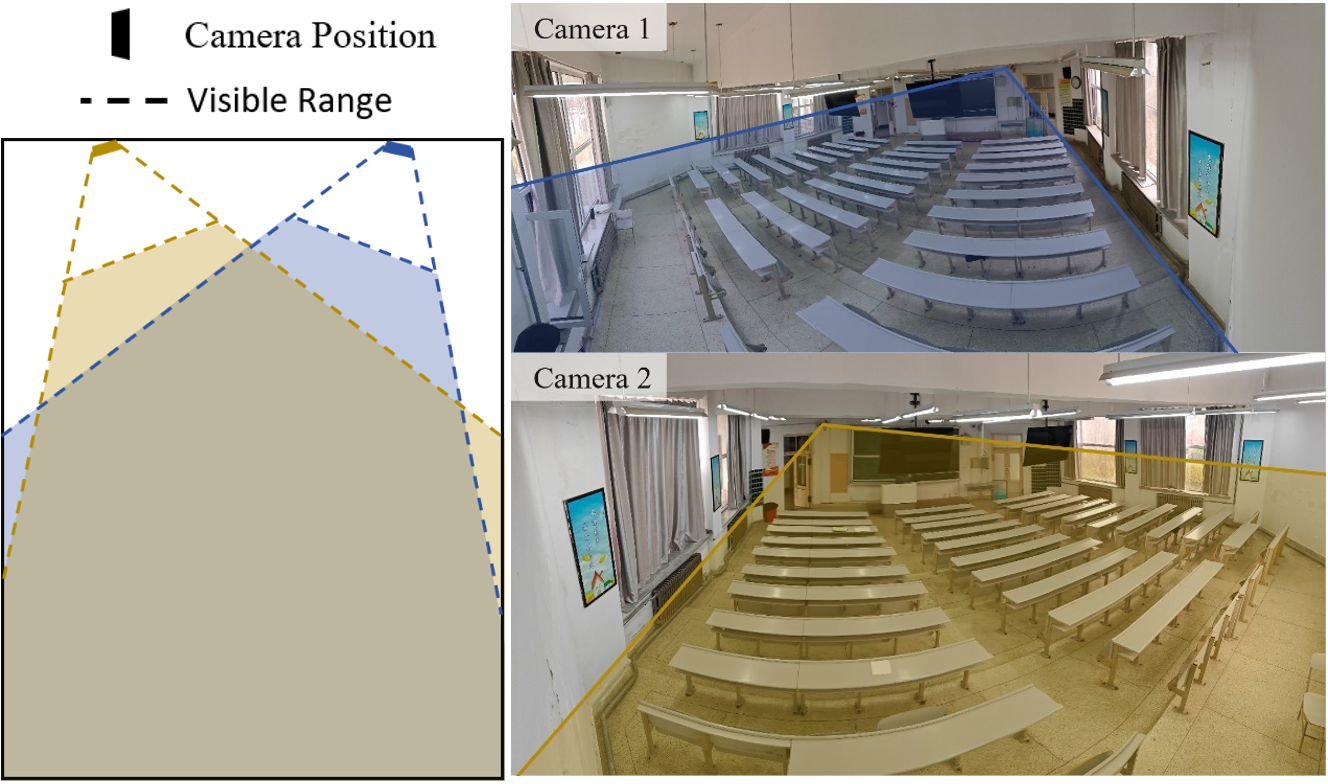}
\caption{Data acquisition environment and dual-view camera deployment of HCCB. The figure shows the relative positions and fields of view of the two cameras. The right side presents real lecture hall scenes captured from the two viewpoints. The blue and yellow regions indicate the primary coverage areas of the two viewpoints, respectively.}
\label{fig:data_acquisition_dual_view}
\end{figure}
After obtaining the raw classroom videos, frames were first extracted at fixed time intervals. Blurred frames, empty-scene frames, and invalid frames were then removed, and near-duplicate sample filtering was conducted to reduce image redundancy. All valid images were standardized to a resolution of $1920 \times 1080$. Finally, HCCB contains 796 high-resolution classroom images and 50,229 bounding boxes for student behavior instances. Although the number of images is relatively limited, HCCB still provides a large instance scale and a high detection burden because each image from a real large-scale classroom contains a large number of student instances.
\subsection{Annotation Protocol and Dataset Split}\label{subsec:annotation_protocol_split}
In highly congested classroom scenarios, the difficulty of student behavior detection comes not only from object localization, but also from semantic overlap among neighboring behavior categories. For example, reading, writing, and bowing head~\citep{yang2023scbdataset3,yang2023studentyolov7} may all appear as head-down postures. Using phone and bowing head~\citep{yang2023yolov7bra} may also share similar head orientations, while looking around and heads up are mainly distinguished by head direction and attention orientation. Without a unified annotation protocol, category labels can easily drift between adjacent semantic categories, thereby affecting the stability of model training and evaluation.

To reduce annotation ambiguity, HCCB adopts an evidence-driven and priority-based mutually exclusive annotation paradigm. Specifically, annotators first determine candidate behavior categories according to visible visual evidence and then perform single-label adjudication based on a fixed priority chain. Each student instance is assigned only one behavior label, and multi-label assignment is not used. A smaller priority value indicates a higher adjudication priority. When an instance satisfies the conditions of multiple categories, the category with the higher priority is selected. For example, if a student shows a head-down posture and clear evidence of phone operation is visible, the instance is annotated as using phone rather than bowing head. If a clear writing action is observed, such as the pen tip touching the paper, the instance is preferentially annotated as writing.

HCCB finally defines seven objectively observable behaviors, including reading, writing, heads up, sleeping, looking around, bowing head, and using phone. Among them, head-down-related behaviors such as reading, writing, and bowing head follow the category settings of existing classroom behavior detection tasks~\citep{yang2023scbdataset3,yang2023studentyolov7}, while abnormal behavior categories such as using phone are consistent with the requirements of classroom state monitoring~\citep{yang2023yolov7bra}. The definitions, instance numbers, and annotation priorities of all categories are shown in Table~1.
\begin{table}[width=.96\linewidth,cols=4,pos=htbp]
\centering
\caption{Category definitions, instance numbers, and annotation priorities in HCCB.}
\label{tab:hccb_behavior_definitions}
\small
\setlength{\tabcolsep}{3pt}
\begin{tabular}{@{}c l r >{\raggedright\arraybackslash}p{11cm}@{}}
\toprule
Priority & Class & Instances & Annotation Criterion \\
\midrule
1 & Sleeping & 201 & The upper body is clearly lying on the desk. \\
2 & Using Phone & 4,395 & The student is operating a mobile phone. \\
3 & Writing & 2,327 & A writing action is visible, with the pen tip touching the paper. \\
4 & Reading & 7,492 & The student looks downward at a reading medium and the paper region is visible. \\
5 & Looking Around & 4,883 & The student keeps the head raised and shows a non-forward attention shift. \\
6 & Heads Up & 14,147 & The student keeps the head raised with a stable forward attention state. \\
7 & Bowing Head & 16,784 & The student keeps the head lowered. \\
\bottomrule
\end{tabular}
\end{table}

From the perspective of semantic design, the seven behaviors in HCCB are not loose descriptions of classroom states, but verifiable detection categories based on visual evidence. Among them, sleeping, using phone, and writing rely on clearer action or object evidence and therefore have higher adjudication priorities. Bowing head is the base category for head-down states and is mainly used to cover head-down instances without evidence of reading, writing, or phone use; therefore, it has the lowest priority. This design reduces annotation conflicts between adjacent categories while preserving fine-grained behavior differences in real classrooms.
\begin{figure}
\centering
\includegraphics[width=.90\linewidth]{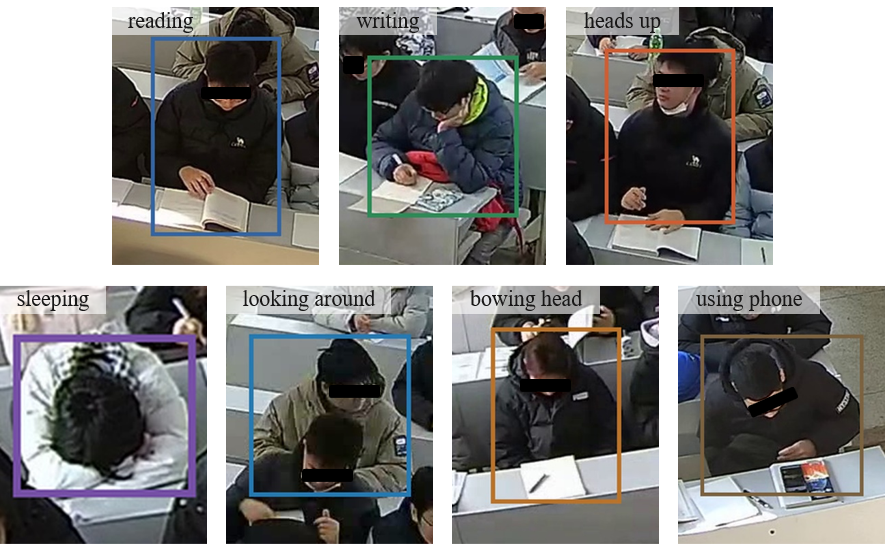}
\caption{Local annotation examples of seven classroom behavior categories in HCCB.}
\label{fig:local_annotation_examples}
\end{figure}

\begin{figure}
\centering
\includegraphics[width=.98\linewidth]{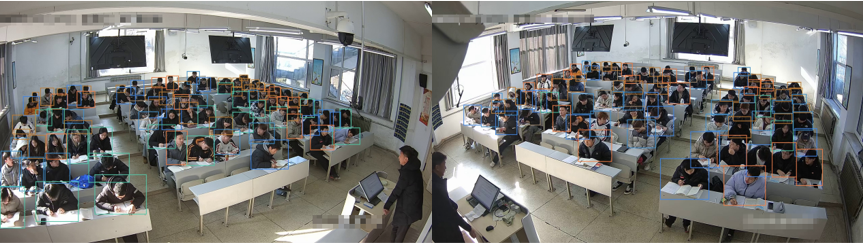}
\caption{Full-image annotation visualization under dual-view classroom scenes.}
\label{fig:full_image_annotation_visualization}
\end{figure}
HCCB is divided into training and validation sets at an approximate ratio of 9:1, with 45,000 instances for training and 5,229 instances for validation. This split maintains image-level independence while preserving class distribution consistency between the training and validation sets as much as possible.

The class-wise distributions of the training and validation sets are reported in Table~2. Overall, HCCB exhibits a clear long-tailed distribution. Bowing Head and Heads Up are the dominant categories, accounting for 33.41\% and 28.17\% of all instances, respectively. In contrast, Sleeping is a markedly low-frequency category, accounting for only 0.40\%. Reading, Looking Around, and Using Phone fall into the middle-frequency range, whereas Writing has a relatively low proportion. This distribution reflects the natural occurrence frequency of student behaviors in real classrooms, but also increases the difficulty of robust learning under class imbalance.

\begin{table}[width=.92\linewidth,cols=4,pos=htbp]
\centering
\caption{Class-wise instance distribution of HCCB in the training and validation sets.}
\label{tab:hccb_class_distribution}
\begin{tabular}{lrrrrr}
\toprule
Class & Train Instances & Val. Instances & Train Ratio & Val. Ratio & Overall Ratio \\
\midrule
Reading & 6,655 & 837 & 14.79\% & 16.01\% & 14.92\% \\
Writing & 2,091 & 236 & 4.65\% & 4.51\% & 4.63\% \\
Heads Up & 12,715 & 1,432 & 28.26\% & 27.39\% & 28.17\% \\
Sleeping & 182 & 19 & 0.40\% & 0.36\% & 0.40\% \\
Looking Around & 4,383 & 500 & 9.74\% & 9.56\% & 9.72\% \\
Bowing Head & 15,026 & 1,758 & 33.39\% & 33.62\% & 33.41\% \\
Using Phone & 3,948 & 447 & 8.77\% & 8.55\% & 8.75\% \\
\midrule
Total & 45,000 & 5,229 & 100.00\% & 100.00\% & 100.00\% \\
\bottomrule
\end{tabular}
\end{table}
As shown in Table~2, the class proportions in the training and validation sets are generally consistent, indicating that the data split effectively preserves class distribution stability. Meanwhile, the class distribution of HCCB exhibits two important characteristics. First, the dominant categories show strong appearance similarity. For example, Heads Up and Looking Around both correspond to head-raised postures, whereas Reading, Writing, and Bowing Head all involve head-down states. This requires the model to perform fine-grained behavior discrimination. Second, the low-frequency category Sleeping has very limited samples and is often accompanied by occlusion and pose variation, which further increases the difficulty of tail-class detection. Therefore, HCCB is not only a high-density classroom detection dataset, but also a challenging benchmark involving class imbalance, semantic confusion among neighboring categories, and degradation of fine-grained visual evidence.

Through the above data collection and annotation pipeline, HCCB preserves high density, severe occlusion, and depth-induced scale variation in real large-scale classrooms while ensuring annotation consistency. Among these characteristics, high density and severe occlusion are core challenges repeatedly emphasized in generic dense crowd detection datasets~\citep{shao2018crowdhuman,zhang2020widerperson}, whereas depth-induced scale variation and fine-grained behavior semantics are structural challenges specific to classroom scenarios.

\subsection{Dataset Difficulty Structure Analysis}\label{subsec:difficulty_structure_analysis}
To further characterize the difficulty of HCCB, we conduct quantitative analysis from three aspects: object density, occlusion burden, and scale-depth structure. Unlike model experiments, all statistics in this section are computed from annotated bounding boxes, category labels, object center positions, and object scales, without relying on any detector outputs. Therefore, these metrics reflect the intrinsic structural difficulty of the dataset.
\subsubsection{Quantitative Metrics of Difficulty Structure}\label{subsubsec:difficulty_metrics}
First, the number of instances per image is used to measure dataset density~\citep{shao2018crowdhuman}. Suppose that the dataset contains \(N\) images, and the \(t\)-th image contains \(n_t\) object instances. The image-level high-density sample ratio is defined as
\begin{equation}\label{eq:p_img}
P_{\mathrm{img}}(\tau)=\frac{1}{N}\sum_{t=1}^{N}\mathbb{I}(n_t\geq \tau),
\end{equation}
where \(\tau\) denotes the instance-number threshold, and \(\mathbb{I}(\cdot)\) is the indicator function. Here, \(P_{\mathrm{img}}(30)\) is used to measure the proportion of medium-to-high-density samples, indicating whether crowding is the dominant pattern in the data distribution. \(P_{\mathrm{img}}(50)\) is used to measure the proportion of extremely high-density samples, reflecting the detection pressure under strong instance competition.

Second, for occlusion burden, since most object detection datasets provide only a single bounding box rather than paired full-body and visible boxes, this paper adopts a reproducible object-level occlusion proxy metric~\citep{shao2018crowdhuman}. For the \(i\)-th object box \(B_i\), its maximum coverage ratio is computed as
\begin{equation}\label{eq:occ_proxy}
o_i=\max_{j\neq i}\frac{|B_i\cap B_j|}{|B_i|},
\end{equation}
where \(B_j\) denotes another object box in the same image, \(|B_i\cap B_j|\) is the intersection area between \(B_i\) and \(B_j\), and \(|B_i|\) is the area of \(B_i\). The object-level occlusion proportion is further defined as
\begin{equation}\label{eq:p_obj}
P_{\mathrm{obj}}(\delta)=\frac{1}{M}\sum_{i=1}^{M}\mathbb{I}(o_i\geq \delta),
\end{equation}
where \(M\) denotes the total number of object boxes and \(\delta\) is the occlusion threshold. Specifically, \(P_{\mathrm{obj}}(0.1)\) measures the proportion of effectively occluded objects among all objects, where the threshold of 0.1 excludes slight contacts or boundary noise. \(P_{\mathrm{obj}}(0.3)\) further characterizes the object coverage burden under stronger occlusion conditions.

Finally, for scale-depth conflict, the normalized vertical coordinate of the object-box center and the normalized object area are used to characterize front-to-rear scale variation in classrooms. Let \(\tilde{y}_i\) denote the normalized vertical coordinate of the center of the \(i\)-th object box, and let \(a_i\) denote its normalized area. The front-to-rear scale ratio is defined as
\begin{equation}\label{eq:front_back_ratio}
R_{\mathrm{F}/\mathrm{B}}
=
\frac{|\mathcal{B}|\sum_{i\in\mathcal{F}} a_i}
{|\mathcal{F}|\sum_{i\in\mathcal{B}} a_i},
\end{equation}
where \(\mathcal{F}\) and \(\mathcal{B}\) denote the front-row and rear-row target sets, respectively. A larger \(R_{\mathrm{F}/\mathrm{B}}\) indicates a more significant scale discontinuity between front-row and rear-row targets. Furthermore, to analyze the distribution pattern of occlusion along the classroom depth direction, the normalized vertical coordinates of object centers are discretized into \(K\) depth bins:
\begin{equation}\label{eq:depth_bin}
\mathcal{D}_k=\left\{i \mid \frac{k-1}{K}\leq \tilde{y}_i < \frac{k}{K}\right\}, \quad k=1,\ldots,K.
\end{equation}
The occlusion proportion in the \(k\)-th depth bin is calculated as
\begin{equation}\label{eq:p_bin}
P_{\mathrm{bin}}^{(k)}(\delta)=\frac{1}{|\mathcal{D}_k|}\sum_{i\in\mathcal{D}_k}\mathbb{I}(o_i\geq \delta).
\end{equation}

Based on the above metrics in Eqs.~\eqref{eq:p_img}--\eqref{eq:p_bin}, this paper compares HCCB with CrowdHuman, CityPersons, SBD, STBD-08, and SCB-D3-S. The results are shown in Table~3.

\begin{table}[width=.94\linewidth,cols=5,pos=htbp]
\centering
\caption{Quantitative comparison of HCCB with representative dense crowd and classroom behavior datasets.}
\label{tab:dataset_difficulty_comparison}
\small
\setlength{\tabcolsep}{4pt}
\begin{tabular}{lrrrrrrr}
\hline
Dataset & Images & Instances & Avg./Img. &
$P_{\mathrm{img}}(30)$ &
$P_{\mathrm{img}}(50)$ &
$P_{\mathrm{obj}}(0.1)$ &
$R_{\mathrm{F}/\mathrm{B}}$ \\
\hline
CrowdHuman & 19,370 & 439,046 & 22.67 & 22.12 & 8.33 & 90.07 & 1.77 \\
CityPersons & 3,475 & 23,593 & 6.79 & 3.31 & 0.72 & 53.61 & 4.45 \\
SBD & 3,773 & 145,338 & 38.52 & 25.15 & 20.09 & \textbf{93.61} & 4.06 \\
STBD-08 & 8,884 & 267,888 & 30.15 & 44.51 & 3.98 & 83.85 & 4.83 \\
SCB-D3-S & 5,015 & 25,806 & 5.15 & 0.74 & 0.00 & 64.28 & 4.95 \\
HCCB (ours) & 796 & 50,229 & \textbf{63.10} & \textbf{95.48} & \textbf{92.96} & 82.20 & \textbf{7.76} \\
\hline
\end{tabular}

\vspace{1mm}
\begin{minipage}{\linewidth}
\footnotesize
Note: $P_{\mathrm{img}}(30)$ and $P_{\mathrm{img}}(50)$ denote the proportions of images containing at least 30 and 50 instances, respectively. $P_{\mathrm{obj}}(0.1)$ denotes the proportion of effectively occluded objects. $R_{\mathrm{F}/\mathrm{B}}$ denotes the front-to-rear scale ratio.
\end{minipage}
\end{table}
As shown in Table~3, HCCB has an average of 63.10 instances per image, which is substantially higher than all compared datasets. Meanwhile, \(P_{\mathrm{img}}(30)\) and \(P_{\mathrm{img}}(50)\) reach 95.48\% and 92.96\%, respectively, indicating that high-density samples are not isolated outliers but dominate the dataset distribution. HCCB also obtains \(P_{\mathrm{obj}}(0.1)=82.20\%\), showing that object-level occlusion is widespread in classroom scenarios. In addition, the front-to-rear scale ratio of HCCB reaches \(R_{\mathrm{F}/\mathrm{B}}=7.76\), which is higher than that of all compared datasets and indicates a pronounced front-to-rear scale discontinuity in multi-row tiered lecture classrooms.

\subsubsection{Structural Visualization}\label{subsubsec:structural_visualization}
Fig.~5 presents the image-level instance density distributions of different datasets. Compared with the other datasets, both the median and interquartile range of HCCB are shifted toward higher values, indicating that each image contains more target instances. Moreover, the entire distribution of HCCB is located in a higher density range, suggesting that its high-density characteristic is not driven by only a small number of highly crowded images.

\begin{figure}
\centering
\includegraphics[width=.95\linewidth]{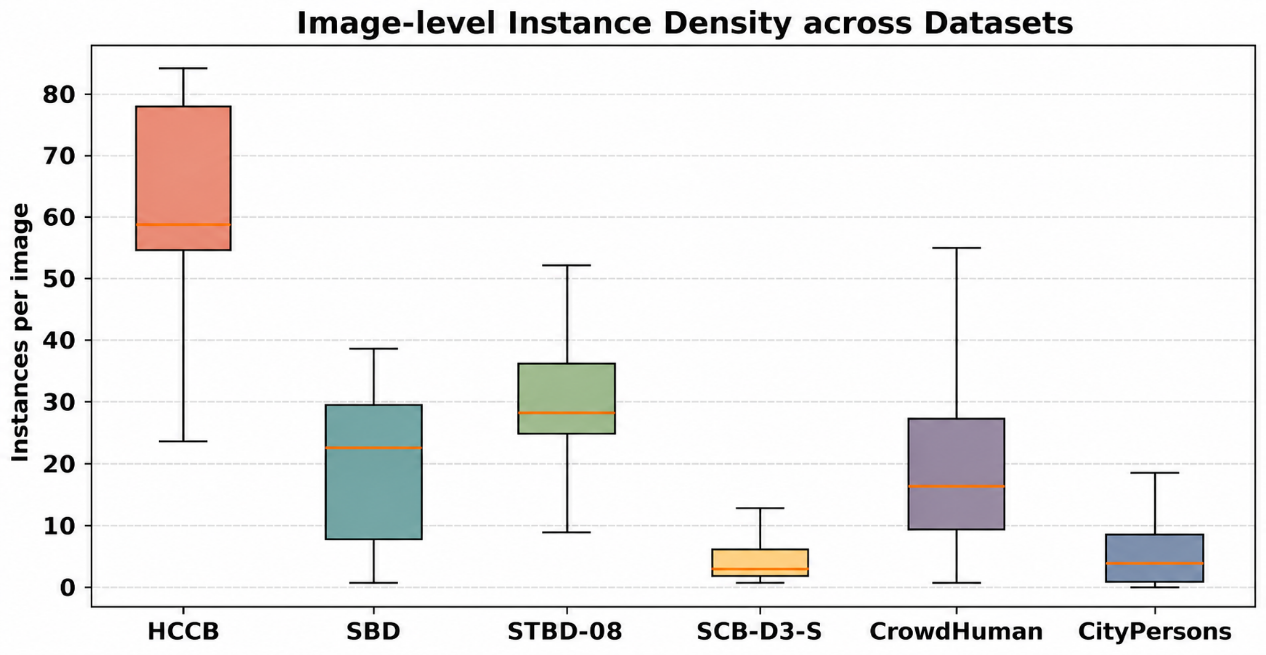}
\caption{Image-level instance density distribution across datasets. The box indicates the interquartile range, the line inside the box indicates the median, and the whiskers indicate the distribution range.}
\label{fig:instance_density_distribution}
\end{figure}
In addition to instance quantity, HCCB exhibits clear spatial distribution characteristics. Fig.~6 presents the normalized spatial distribution of the center points of all student targets. The target centers are mainly distributed in the seating area and form relatively continuous band-like and array-like structures along the classroom seating layout. This distribution is closely related to fixed seats, camera viewpoints, and the front-to-rear spatial arrangement of the classroom.

\begin{figure}
\centering
\includegraphics[width=.78\linewidth]{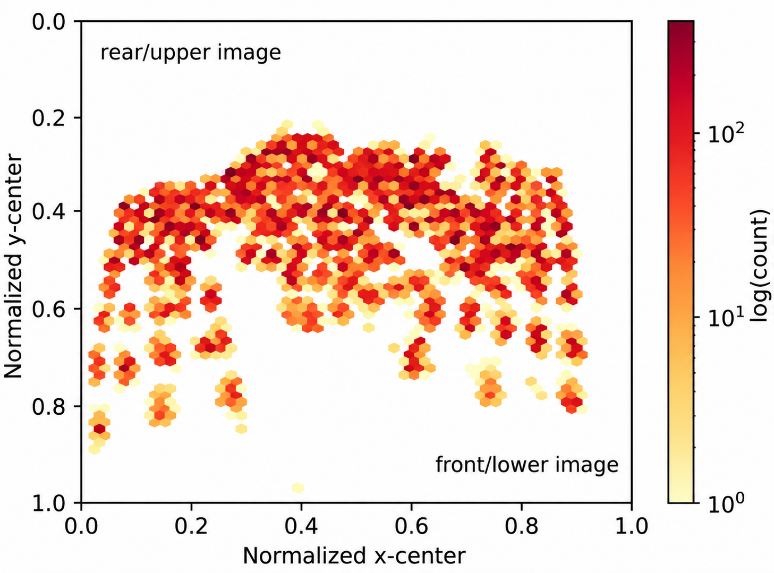}
\caption{Spatial heatmap of object centers in HCCB. The x- and y-axes denote the normalized coordinates of bounding-box centers.}
\label{fig:object_center_heatmap}
\end{figure}

Fig.~7 shows the proportions of mild and severe occlusion under different depth bins. Compared with the other datasets, both the mild-occlusion and severe-occlusion curves of HCCB show clear peaks in the middle and rear regions, indicating that occlusion is not uniformly random but closely associated with the classroom depth structure. Since front-row targets are larger and rear-row targets are smaller, front-row students, desk-chair edges, and partial limbs are more likely to cover the head, shoulder, hand, or associated object regions of rear-row students, resulting in typical asymmetric occlusion.
\begin{figure}
\centering
\begin{minipage}{0.92\linewidth}
\centering
\includegraphics[width=\linewidth]{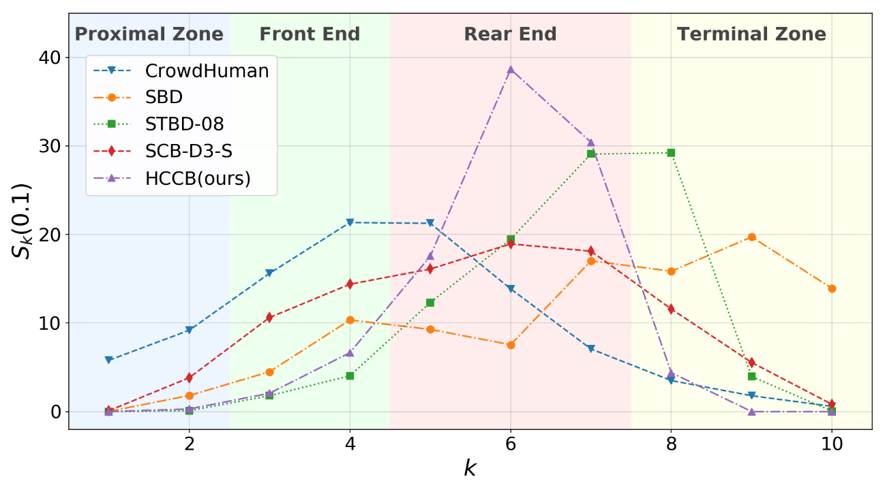}
\vspace{1mm}
\small (a) Light occlusion ratio curve
\end{minipage}
\vspace{2mm}

\begin{minipage}{0.92\linewidth}
\centering
\includegraphics[width=\linewidth]{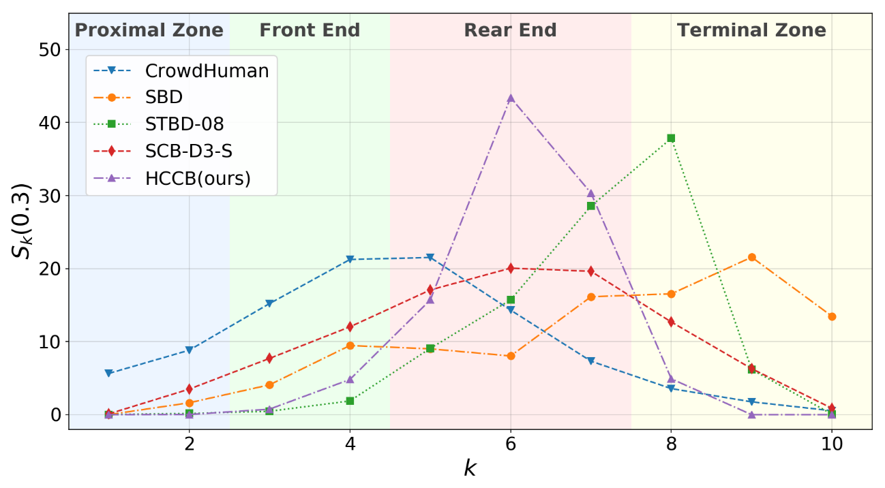}
\vspace{1mm}
\small (b) Heavy occlusion ratio curve
\end{minipage}
\caption{Depth-binned occlusion ratio curves. The horizontal axis \(k\) represents the depth-bin index from the front rows to the back rows. The vertical axis \(P_{\mathrm{bin}}^{(k)}(\delta)\) denotes the occlusion proportion in the corresponding depth region under threshold \(\delta\).}
\label{fig:depth_binned_occlusion_ratio}
\end{figure}
To further describe the relationship between target scale and imaging depth, Fig.~8 presents the variation of object-box area with the normalized vertical center coordinate in different datasets. In HCCB, the object-box area shows a consistent increasing trend as the vertical coordinate increases, indicating that front-row targets occupy larger image regions. In contrast, the scale variation curves of the other datasets fluctuate more substantially, and their overall trends are less continuous than that of HCCB. This phenomenon is related to the spatial distance difference between front and rear rows in tiered lecture classrooms and the fixed camera viewpoint, reflecting a relatively stable depth-scale gradient in HCCB.
\begin{figure}
\centering
\includegraphics[width=.92\linewidth]{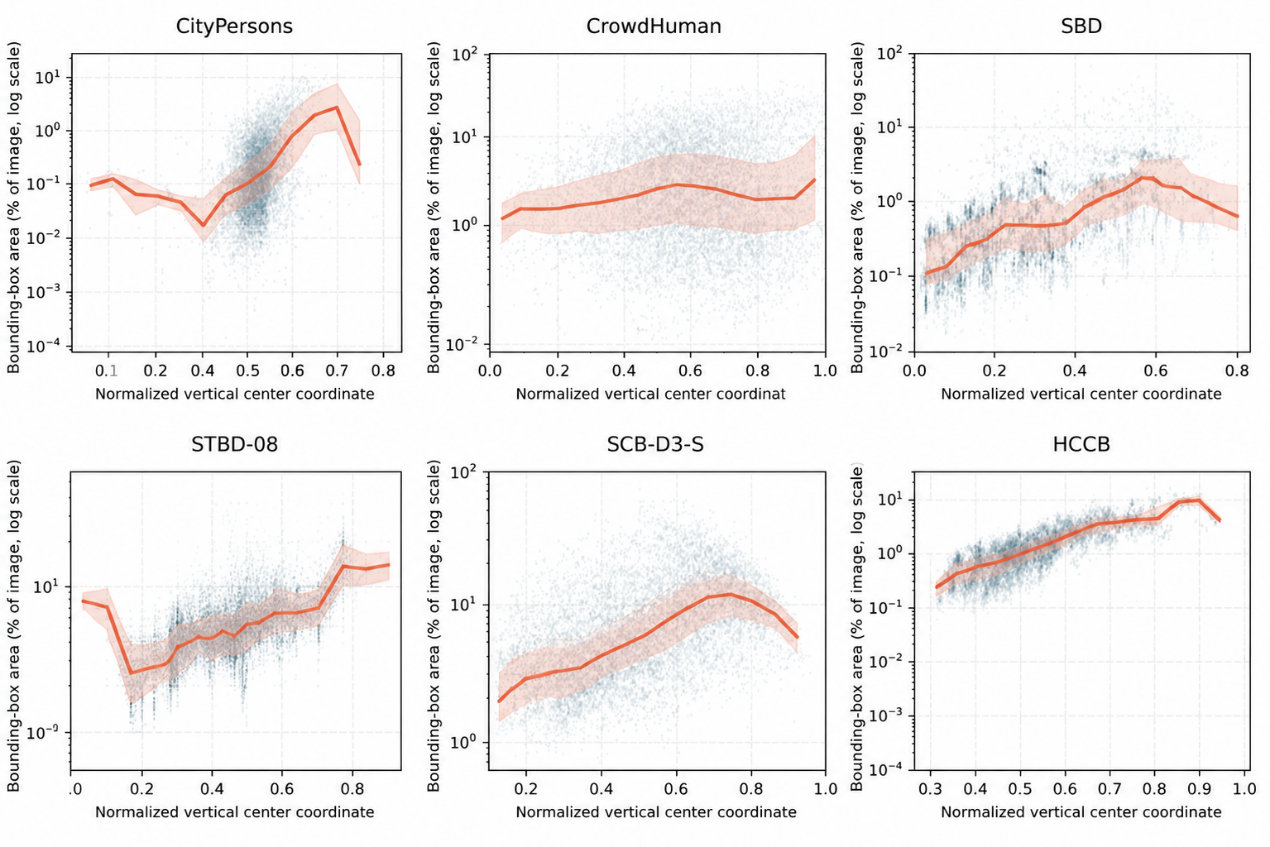}
\caption{Cross-dataset relationship between object scale and vertical position. Points represent individual object instances, the curve shows the binned average trend, and the shaded region indicates the corresponding statistical range.}
\label{fig:scale_vertical_position_relationship}
\end{figure}
In addition, the behavior-category distribution of HCCB is related to classroom depth regions. Fig.~9 presents the relative proportions of different behavior categories in the rear, middle, and front regions. From rear to front regions, the proportions of Reading and Writing increase, whereas the proportion of Bowing Head decreases. Since categories such as Reading, Writing, Bowing Head, and Using Phone rely on local visual cues, including paper regions, hands, phone regions, or head posture, the smaller target scale and stronger occlusion in the middle and rear regions further increase the difficulty of behavior recognition.
\begin{figure}
\centering
\includegraphics[
    width=.88\linewidth,
    height=.43\linewidth
]{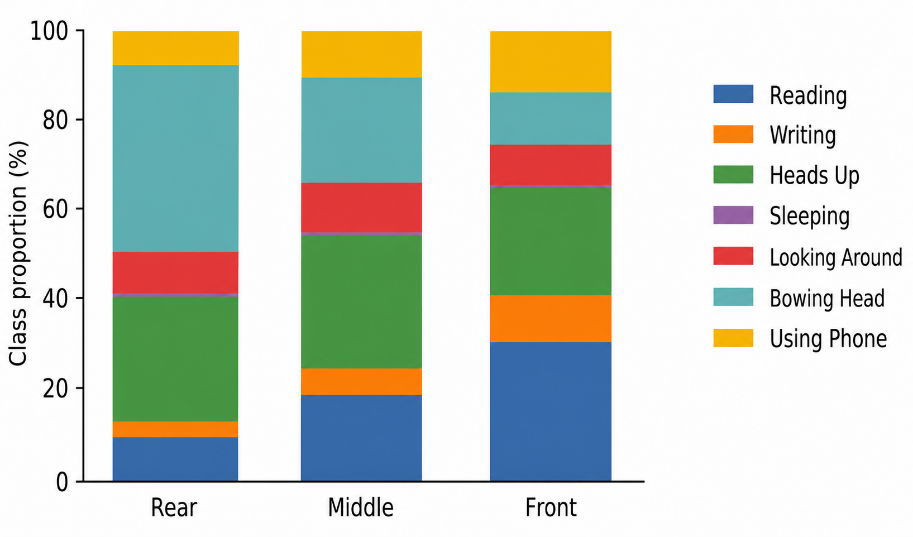}
\caption{Behavior-category distribution across different depth regions in HCCB.}
\label{fig:depth_region_behavior_distribution}
\end{figure}

\subsubsection{Detection Bottlenecks Induced by the Dataset}\label{subsubsec:dataset_induced_bottlenecks}
Based on the above statistical results, the difficulty of HCCB can be summarized into four detection bottlenecks. First, HCCB contains dense concurrent instances. Each image includes a large number of student instances, and the target centers form a stable dense array in the seating area, which can easily lead to foreground response competition, box merging, and positive sample assignment congestion~\citep{zhang2020atss,ge2021ota}. Second, HCCB exhibits asymmetric occlusion. The occlusion peaks are concentrated in the middle and rear regions, indicating that the camera viewpoint, large front-row targets, and desk-chair structures jointly cause persistent occlusion of small rear-row targets. As a result, rear-row instances retain only limited visible evidence. Third, HCCB presents depth-induced scale discontinuity. Target scale is highly correlated with the front-to-rear classroom position. Large front-row targets and small rear-row targets appear simultaneously, increasing the difficulty of cross-scale feature fusion~\citep{lin2017fpn,tan2020efficientdet} and semantic alignment between shallow and deep features. Fourth, HCCB suffers from far-field semantic degradation. Middle- and rear-row targets are usually smaller, blurrier, and more easily occluded. Meanwhile, categories such as Reading, Writing, Bowing Head, and Using Phone strongly depend on local details. Therefore, their semantic boundaries are more easily compressed.

Therefore, the challenge of HCCB does not arise from the increase of a single statistic, but from the coupled effects of dense concurrent instances, asymmetric occlusion, depth-induced scale discontinuity, and far-field semantic degradation. Based on these dataset characteristics, this paper proposes ODER-HSFNet in Section~4 to improve the robustness of highly congested classroom behavior detection from three aspects: occlusion-aware visible evidence compensation, cross-scale high-order relation modeling~\citep{feng2019hypergraph,kipf2017gcn}, and detection-side candidate-box calibration~\citep{bodla2017softnms,li2020gfl}.

\section{Methodology and Design}\label{sec:methodology_design}

\subsection{System Overview}\label{subsec:system_overview}
The detection difficulty of HCCB is not a simple performance degradation caused by an increased number of objects, but arises from structural failure modes jointly induced by extremely high instance density, asymmetric occlusion, depth-induced scale discontinuity, and far-field semantic degradation. Dense instances cause local response competition and candidate-box congestion~\citep{zhang2020atss,li2020gfl}. The occlusion of rear-row targets by front-row targets weakens the visible evidence of small objects. Perspective variation in multi-row tiered lecture classrooms aggravates cross-scale feature mismatch. The combination of low resolution and occlusion in distant regions further compresses the semantic boundaries among fine-grained behavior categories such as Reading, Writing, Bowing Head, and Using Phone. To address these challenges, this paper constructs ODER-HSFNet based on the YOLO framework~\citep{redmon2016yolo,lei2025yolov13}, and its overall architecture is shown in Fig.~10.
\begin{figure}
\centering
\includegraphics[width=.76\linewidth]{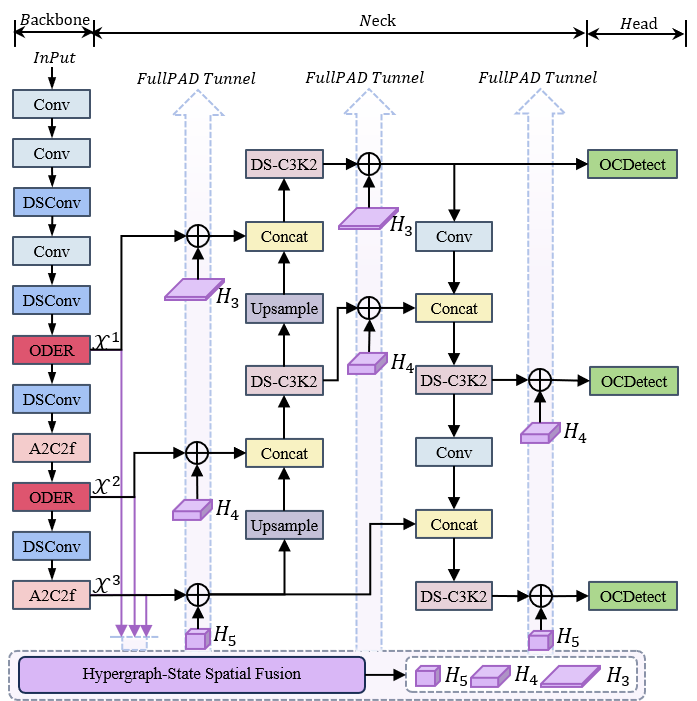}
\caption{Overall framework of ODER-HSFNet.}
\label{fig:oder_hsfnet_framework}
\end{figure}
ODER-HSFNet is a YOLO-based detection framework for highly congested classroom behavior detection and follows the backbone-neck-head paradigm of YOLO~\citep{redmon2016yolo}. The input image is first processed by the YOLO Backbone to extract multi-level features. Then, ODER is introduced into the \(X^1\) and \(X^2\) feature layers to perform controlled compensation of reliable visible evidence around occlusion edges and far-field degraded regions. The multi-scale features refined by ODER are further fed into HSSF to perform cross-scale state modeling, high-order relation fusion, and local detail preservation. Finally, the enhanced features pass through the YOLO Neck and OCDetect to generate the final detection results.

ODER is designed to address visible evidence loss under asymmetric occlusion. Conventional convolution uses a fixed sampling grid, which easily mixes responses from target regions, occluded regions, and neighboring instances. Although unconstrained deformable sampling can dynamically adjust the receptive field, it may cross into neighboring students or background regions in highly congested classrooms. ODER re-aggregates reliable visible evidence within a controlled neighborhood through bounded deformable edge resampling~\citep{dai2017deformable}, topology-aware sampling routing, and sample-level residual amplitude modulation. This design reduces feature contamination around occlusion edges and improves the representation stability of distant small targets.

HSSF is designed to address cross-layer feature mismatch caused by high-density instance competition and depth-induced scale discontinuity. It takes three-scale features, namely \(X^1\), \(X^2\), and \(X^3\), as inputs. First, DSConv~\citep{howard2017mobilenets} and VSS Proxy~\citep{gu2023mamba,liu2024vmamba} are used to enhance local structural cues and long-range contextual dependencies. Then, scale-aligned fusion is performed under a unified spatial reference. Finally, hypergraph semantic relations among dense student instances are modeled through a high-order relation fusion bottleneck~\citep{feng2019hypergraph}. By integrating local structure preservation, state-space propagation, and high-order relation aggregation, HSSF alleviates the smoothing of small-object details, semantic mismatch between shallow and deep features, and dense response competition in traditional pyramid fusion~\citep{lin2017fpn,tan2020efficientdet}.

OCDetect is designed to address candidate-box noise and confidence distortion at the detection output stage. In highly congested classrooms, desk-chair edges, partial limbs, occlusion boundaries, and background textures can generate a large number of low-quality candidate boxes. In addition to the original classification and regression branches, OCDetect introduces a class-agnostic objectness calibration branch and performs denoising correction on category confidence before NMS~\citep{bodla2017softnms,liu2019adaptive}. This design does not change the bounding-box regression path, but improves candidate-box ranking and post-processing stability in dense scenes by suppressing false responses induced by background noise and occlusion boundaries.

From the perspective of the correspondence between structural difficulties and model components, ODER restores effective evidence around occlusion edges and far-field degraded regions at the local sampling level, HSSF enhances cross-scale high-order relation modeling at the feature fusion level, and OCDetect suppresses noisy Pre-NMS candidate boxes at the detection output level. Together, these three modules form a complete enhancement path of local evidence compensation, cross-scale high-order fusion, and detection-side confidence calibration, enabling ODER-HSFNet to more stably adapt to highly dense and occluded behavior detection tasks in real large-scale classrooms.
\subsection{Occlusion-aware Deformable Edge Rectifier}\label{subsec:oder}
In HCCB, asymmetric occlusion and far-field degradation are key factors that degrade local discriminative evidence. Front-row students, desk-chair edges, and adjacent body parts often occlude rear-row small targets, leaving only partial visible regions such as the head, shoulders, hands, or associated objects. Fixed convolutional sampling easily mixes responses from the target, the occluder, and neighboring instances. Although unconstrained deformable sampling~\citep{dai2017deformable,zhu2021deformabledetr} can adaptively adjust the receptive field, it may introduce cross-instance responses among highly dense neighboring targets. To address this problem, this paper designs the Occlusion-aware Deformable Edge Rectifier (ODER), which performs feature resampling and evidence compensation for occlusion edges and far-field degraded regions within a controlled neighborhood.
\begin{figure}
\centering
\includegraphics[width=.92\linewidth]{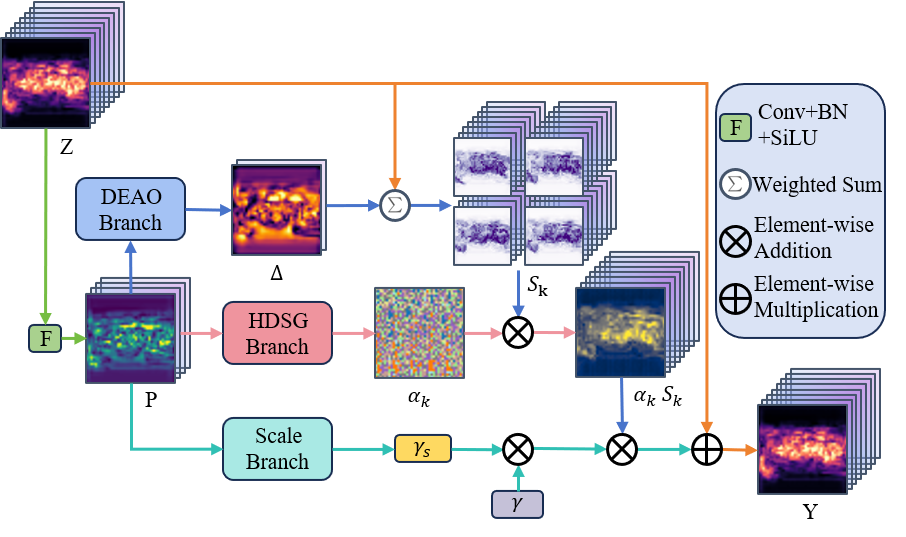}
\caption{Structure of the ODER module.}
\label{fig:oder_module_structure}
\end{figure}
The structure of ODER is shown in Fig.~11. Given an input feature \(X\), ODER first performs channel alignment through convolution to obtain the main-path feature \(Z\), and then generates the shared prior feature \(P\). This process is formulated as
\begin{equation}
Z=\mathrm{SiLU}\left(W_{\mathrm{in}}(X)\right),
\end{equation}
\begin{equation}
P=\phi_p(Z),
\end{equation}
where \(W_{\mathrm{in}}\) denotes the input channel alignment mapping, \(\phi_p(\cdot)\) denotes the prior encoding mapping, and SiLU denotes the nonlinear activation function. The shared prior \(P\) is simultaneously used for offset prediction, sampling-weight estimation, and residual amplitude modulation, enabling different branches to model local structures based on consistent contextual information.
\begin{figure}
\centering
\includegraphics[width=.92\linewidth]{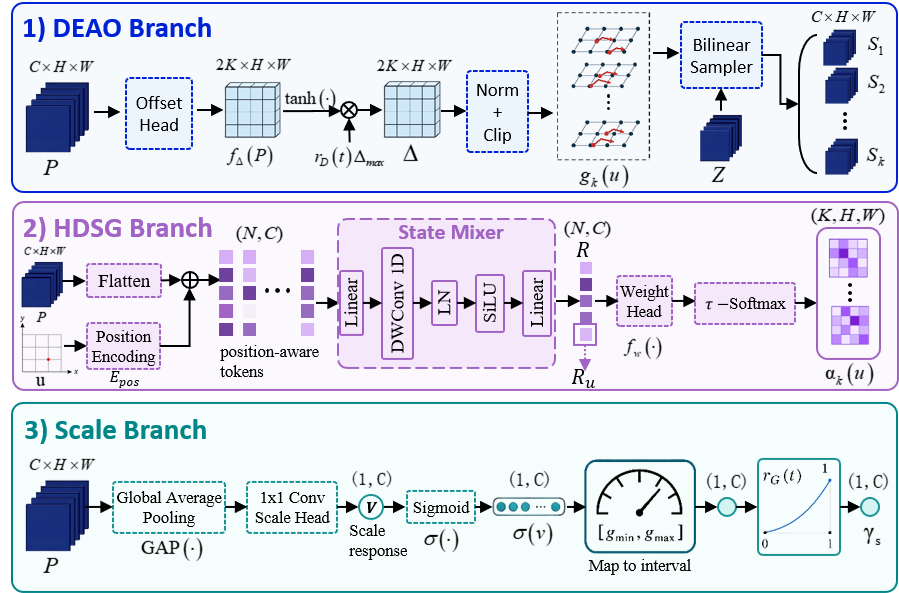}
\caption{Detailed three-branch structure of the ODER module. The figure shows the internal computation flows of the DEAO, HDSG, and Scale branches, including offset generation, sampling-weight prediction, and residual-amplitude modulation.}
\label{fig:oder_three_branch_structure}
\end{figure}
As shown in Fig.~12, ODER consists of three branches: DEAO, HDSG, and Scale. The DEAO branch predicts multiple groups of two-dimensional sampling offsets from the shared prior \(P\), and obtains candidate resampled features through bounded constraints, normalized clipping, and bilinear sampling. The HDSG branch unfolds spatial features into position-aware tokens, models spatial context through a state mixer~\citep{gu2023mamba,liu2024vmamba}, and generates normalized routing weights for different sampling hypotheses. The Scale branch estimates a sample-level residual modulation factor from global context to control the amplitude of compensatory features injected into the main path. These three branches correspond to sampling-evidence generation, sampling-evidence selection, and residual-amplitude control, respectively, and jointly form the internal computation flow of ODER.

For bounded deformable edge resampling, DEAO first predicts \(K\) groups of two-dimensional offsets from the shared prior feature \(P\). To prevent sampling points from drifting drastically in the early training stage, the offset magnitude is constrained using the \(\tanh\) function and a warmup coefficient. Here, \(f_{\Delta}(\cdot)\) denotes the offset prediction mapping, \(\Delta_{\max}\) denotes the maximum offset magnitude, and \(r_D(t)\in[0,1]\) is a warmup coefficient that increases with the training step. For a spatial position \(u\), the normalized sampling coordinate of the \(K\)-th sampling hypothesis is obtained by adding the normalized offset to the base grid coordinate and then clipping it to the valid sampling range. The \(K\)-th candidate evidence is then obtained through bilinear grid sampling. This design enables ODER to re-acquire effective evidence near occlusion edges within a limited neighborhood, rather than performing unconstrained large-range searching.

For topology-aware sampling routing, the reliability of different sampling hypotheses is not the same. In occluded regions, some sampling points may fall on visible target edges, whereas others may fall on occluders or background regions. Therefore, ODER further uses topology-aware routing to assign spatial weights to different sampling results. Specifically, the shared prior \(P\) is flattened into spatial tokens, combined with two-dimensional positional encoding, and then fed into a lightweight state mixer to obtain the topology-context representation \(R_u\) at position \(u\). A sampling-weight prediction head then maps \(R_u\) to routing scores for \(K\) sampling hypotheses, followed by temperature-scaled softmax normalization. The obtained weight \(\alpha_k(u)\) represents the contribution of the \(K\)-th sampling hypothesis at position \(u\), and the sum of all sampling weights equals one. Finally, ODER performs weighted aggregation between the topology routing weights and the multi-hypothesis sampling evidence generated by DEAO. Through this routing mechanism, ODER increases the contribution of sampling results along visible target edges and reduces the influence of sampling results from occluders, background regions, or neighboring instances. In the final configuration, this paper adopts a bounded resampling setting with \(K=8\), achieving a balance between sampling diversity and spatial stability. The above formulas retain the general \(K\)-hypothesis form to describe the extensible multi-hypothesis routing mechanism of ODER.

For sample-level residual amplitude modulation, compensatory features should not be injected into the main path with a fixed intensity because different images may exhibit different degrees of occlusion and target density. Therefore, ODER introduces a sample-level amplitude modulation branch to predict the residual injection strength according to global context. Specifically, global average pooling is applied to the shared prior \(P\), and the resulting global representation is mapped to a bounded scale factor through a scale prediction head and a sigmoid activation. A warmup coefficient is further used to gradually release the residual compensation strength during training. Finally, the ODER output is obtained by adding the modulated compensatory evidence to the main-path feature. This residual form ensures that ODER does not directly replace the original features, but injects controlled occlusion-compensated evidence into the main path.

To ensure training stability, ODER adopts stable initialization for the offset prediction head, output mapping head, and scale modulation head, combined with offset warmup and residual-amplitude warmup. This makes the module close to an identity mapping in the early training stage and then gradually releases the capabilities of deformable sampling and feature compensation. In summary, ODER improves the utilization of effective evidence around occlusion edges and far-field degraded regions through bounded edge resampling, topology-aware routing, and sample-level residual modulation, without disrupting the stability of backbone features.

\subsection{Hypergraph-State Spatial Fusion}\label{subsec:hssf}
In highly congested classroom behavior detection, student instances are densely distributed along the classroom depth direction. A single image simultaneously contains large front-row targets, small rear-row targets, and many partially occluded targets. Conventional FPN~\citep{lin2017fpn} and PAN~\citep{liu2018panet} mainly rely on progressive fusion between adjacent scales, which can easily smooth out shallow details and makes it difficult to explicitly model high-order semantic relations among dense instances. To address this issue, this paper proposes the Hypergraph-State Spatial Fusion (HSSF) module. This module takes three-scale features as inputs and sequentially performs cross-scale state proxy construction, scale-aligned fusion, and high-order relation reconstruction, thereby enhancing the model's representation ability for small-object details, depth-induced scale differences, and dense instance relations.

\begin{figure}
\centering
\includegraphics[width=.96\linewidth]{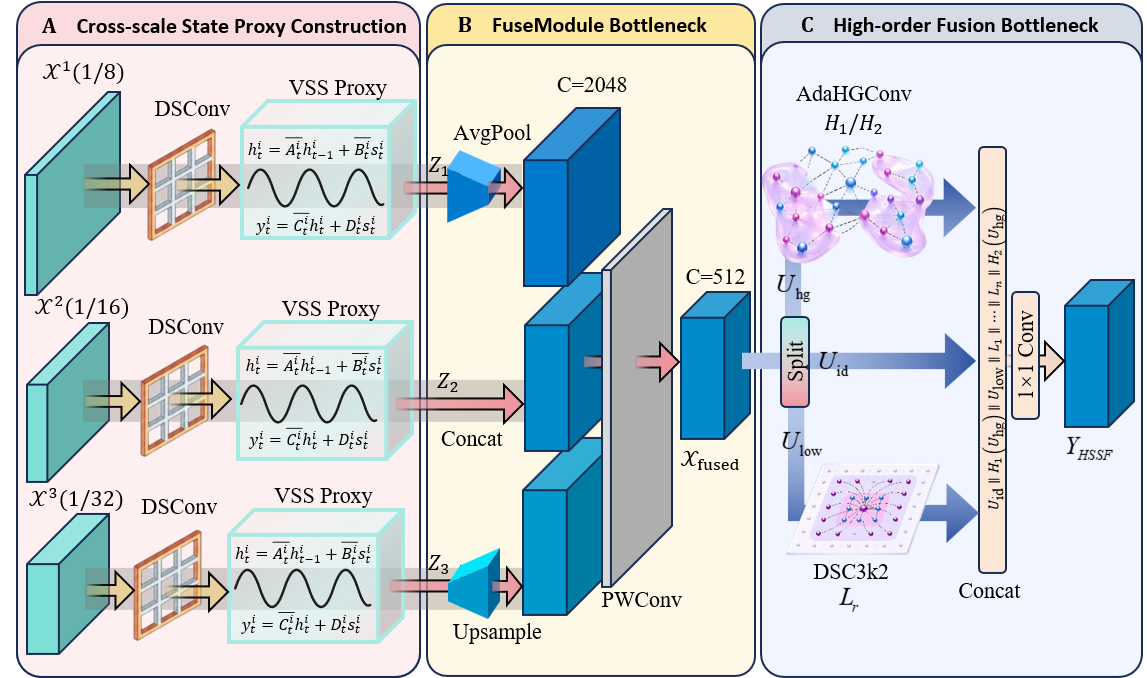}
\caption{Overall architecture of the Hypergraph-State Spatial Fusion (HSSF) mechanism. (A) Cross-scale state proxy construction. (B) FuseModule bottleneck. (C) High-order relation fusion bottleneck.}
\label{fig:hssf_architecture}
\end{figure}
Let the input three-scale features be \(X^1\), \(X^2\), and \(X^3\). As shown in Fig.~13, HSSF consists of three consecutive stages. First, in the cross-scale state proxy construction stage, the three-scale input features are processed by DSConv and VSS Proxy, respectively, to obtain scale proxy features containing local structures and long-range context. Then, in the scale-aligned fusion bottleneck, proxy features at different scales are aligned into a unified representation space through pooling, upsampling, and concatenation, and a fused basis \(X_{\mathrm{f}}\) is generated through \(1\times1\) convolution. Finally, in the high-order relation fusion bottleneck, the fused feature is divided into a high-order relation flow, an identity-preserving flow, and a low-order local flow. After concatenation, the output feature \(Y_{\mathrm{HSSF}}\) is generated through \(1\times1\) convolution.

\paragraph{Cross-scale State Proxy Construction.}
Directly flattening two-dimensional features into sequences may weaken local spatial continuity. Therefore, HSSF first introduces DSConv at each scale for lightweight local structure enhancement, and then uses VSS Proxy to establish long-range contextual dependencies:
\begin{equation}
s_{i,t}=\operatorname{Flatten}\left(\Phi_i^{\mathrm{DS}}\left(X^i\right)\right)_t,
\quad i\in\{1,2,3\}.
\end{equation}
\begin{equation}
\begin{aligned}
h_{i,t} &= A_{i,t}h_{i,t-1}+B_{i,t}s_{i,t},\\ 
z_{i,t} &= C_{i,t}h_{i,t}+D_{i,t}s_{i,t}. 
\end{aligned}
\end{equation}
where \(\Phi_i^{\mathrm{DS}}\) denotes DSConv at the \(i\)-th scale, \(s_{i,t}\) denotes the input token at sequence position \(t\), \(h_{i,t}\) denotes the hidden state variable, and \(z_{i,t}\) denotes the state proxy output. After all \(z_{i,t}\) are restored into spatial features, the enhanced three-scale state features \(Z_i\) are obtained. This process enables HSSF to preserve both local edge cues and long-range contextual information before cross-scale fusion.

\paragraph{Scale-aligned Fusion Bottleneck.}
After state proxy construction, HSSF aligns the three-scale features to the resolution of \(X^2\). Specifically, the high-resolution feature \(Z_1\) is downsampled, the low-resolution feature \(Z_3\) is upsampled, and they are concatenated with \(Z_2\) along the channel dimension. A \(1\times1\) convolution is then used to generate the fused basis:
\begin{equation}
X_{\mathrm{f}}=
\delta\left(
W_f\left[
\mathcal{P}_2\left(Z_1\right)
\parallel Z_2
\parallel
\mathcal{U}_2\left(Z_3\right)
\right]
\right),
\end{equation}
where \(\mathcal{P}_2(\cdot)\) denotes \(2\times\) downsampling, \(\mathcal{U}_2(\cdot)\) denotes \(2\times\) upsampling, \(\parallel\) denotes channel concatenation, \(W_f\) denotes the channel fusion mapping, and \(\delta(\cdot)\) denotes normalization and nonlinear activation after convolution. Unlike progressive pyramid propagation~\citep{lin2017fpn,liu2018panet}, this bottleneck directly fuses shallow details, middle-level structures, and deep semantics under a unified spatial reference, which helps alleviate cross-layer feature mismatch caused by front-to-rear scale discontinuities.

\paragraph{High-order Relation Fusion Bottleneck.}
After obtaining the fused basis \(X_{\mathrm{f}}\), HSSF further performs high-order relation reconstruction. First, \(X_{\mathrm{f}}\) is divided into an identity-preserving flow, a high-order relation flow, and a low-order local flow through channel mapping:
\begin{equation}
\left[
U_{\mathrm{id}},
U_{\mathrm{hg}},
U_{\mathrm{low}}
\right]
=
\operatorname{Split}\left(W_s X_{\mathrm{f}}\right),
\end{equation}
where \(U_{\mathrm{id}}\) is used to preserve the original fused response, \(U_{\mathrm{hg}}\) is used for adaptive hypergraph modeling, and \(U_{\mathrm{low}}\) is used to supplement local textures and edge details.

For the high-order relation flow \(U_{\mathrm{hg}}\), it is flattened into a spatial node sequence \(T\in\mathbb{R}^{B\times N\times c}\), where \(B\) is the batch size, \(H\) and \(W\) are the height and width of the feature map, \(N=HW\) is the number of spatial nodes, and \(c\) is the channel dimension. Sample-specific hyperedge prototypes are generated according to global context:
\begin{equation}
g=
\left[
\operatorname{Mean}_{N}(T)
\parallel
\operatorname{Max}_{N}(T)
\right],
\quad
E=
E_0+
\operatorname{Reshape}\left(W_g g\right),
\end{equation}
where \(g\) denotes the sample-level global descriptor, \(E_0\in\mathbb{R}^{N_h\times d}\) denotes the learnable base hyperedge prototypes, \(E\in\mathbb{R}^{B\times N_h\times d}\) denotes the dynamic hyperedge prototypes of the current sample, \(N_h\) is the number of hyperedges, \(d\) is the hyperedge embedding dimension, \(W_g\) is a learnable projection matrix, and \(\parallel\) denotes feature concatenation. The soft association matrix between nodes and hyperedges is defined as

\begin{equation}\label{eq:hssf_association}
A_{n,e}
=
\frac{
\exp\left(\left(W_qT_n\right)E_e^{\top}/\sqrt{d}\right)
}{
\sum_{n'=1}^{N}
\exp\left(\left(W_qT_{n'}\right)E_e^{\top}/\sqrt{d}\right)
},
\quad n=1,\ldots,N,\ e=1,\ldots,N_h .
\end{equation}

where \(A\in\mathbb{R}^{N\times N_h}\) denotes the node-hyperedge association matrix, \(W_q\) is a learnable query projection, and the normalization is performed over the node dimension for each hyperedge.Based on the association matrix \(A\), adaptive hypergraph convolution~\citep{feng2019hypergraph} models high-order relations through two-stage message passing from nodes to hyperedges and from hyperedges to nodes:

\begin{equation}
H(T)=\phi_v\left(A\phi_e\left(A^{\top}T\right)\right)+T,
\end{equation}
where \(\phi_e(\cdot)\) and \(\phi_v(\cdot)\) denote hyperedge feature transformation and node feature transformation, respectively. A hyperedge in the hypergraph can simultaneously connect multiple spatial nodes, making it more suitable for describing multi-instance associations within the same depth region, behavior pattern, or occlusion structure.

Meanwhile, to prevent the high-order branch from being overly biased toward global semantics, HSSF introduces a lightweight low-order local branch on \(U_{\mathrm{low}}\):

\begin{equation}\label{eq:hssf_low_order}
\begin{aligned}
L_0 &= U_{\mathrm{low}},\\
L_r &= D_r\left(L_{r-1}\right), \quad r=1,\ldots,n,
\end{aligned}
\end{equation}
where \(D_r(\cdot)\) denotes the \(r\)-th lightweight local transformation in the low-order branch, and \(n\) is the number of local transformation blocks.
Finally, the identity-preserving flow, high-order hypergraph flow, and low-order local flow are concatenated, and the HSSF fused feature is obtained through the output mapping:
\begin{equation}
Y_{\mathrm{HSSF}}=
\psi_o\left(
\left[
U_{\mathrm{id}}
\mathbin{\parallel} H_1\left(U_{\mathrm{hg}}\right)
\mathbin{\parallel} U_{\mathrm{low}}
\mathbin{\parallel} L_1
\mathbin{\parallel} \cdots
\mathbin{\parallel} L_n
\mathbin{\parallel} H_2\left(U_{\mathrm{hg}}\right)
\right]
\right),
\end{equation}
where \(H_1(\cdot)\) and \(H_2(\cdot)\) denote two parallel high-order hypergraph branches, and \(\psi_o(\cdot)\) denotes the output \(1\times1\) aggregation mapping. Inspired by the feature reinjection idea of FullPAD in YOLO~\citep{lei2025yolov13}, this paper reinjects \(Y_{\mathrm{HSSF}}\) as the enhanced fused feature into the subsequent detection neck to maintain the overall network connectivity and training stability.

In summary, HSSF integrates local structure, long-range context~\citep{gu2023mamba,liu2024vmamba}, and high-order instance relations~\citep{feng2019hypergraph} into the multi-scale feature fusion process through cross-scale state proxy construction, a scale-aligned fusion bottleneck, and a high-order relation fusion bottleneck. This mechanism can alleviate small-object detail dilution, depth-induced scale mismatch, and dense instance response competition in highly congested classroom scenarios, providing more robust behavior representations for the subsequent detection head.

\subsection{Occlusion-Calibrated Detection Head}\label{subsec:ocdetect}
In highly congested classroom scenarios, the detection head needs not only to perform object classification and bounding-box regression, but also to suppress false responses caused by desk-chair edges, partial limbs, occlusion boundaries, and background textures before candidate boxes enter NMS. During the training and validation of the baseline YOLO, we observe that extremely high-density images tend to generate a large number of low-quality background candidate boxes~\citep{shao2018crowdhuman,li2020gfl}. These boxes increase the burden of NMS post-processing and may lead to unstable candidate-box filtering. To address this issue, this paper proposes OCDetect, which introduces a class-agnostic objectness calibration branch~\citep{redmon2016yolo} into the detection head to enhance background-noise suppression in highly dense and occluded scenarios.
\begin{figure}
\centering
\includegraphics[width=.98\linewidth]{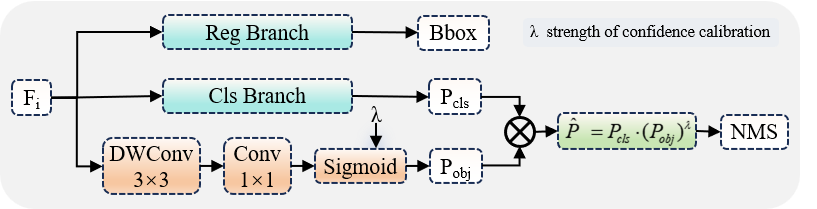}
\caption{Illustration of the Occlusion-Calibrated Detection Head (OCDetect).}
\label{fig:ocdetect_head}
\end{figure}

The structure of OCDetect is shown in Fig.~14. Given the detection feature \(F_i\) at the \(i\)-th scale, the original detection head contains a bounding-box regression branch and a class prediction branch, which output bounding-box predictions and class probabilities, respectively. On this basis, OCDetect additionally introduces a lightweight objectness branch:
\begin{equation}
O_i=
\operatorname{Conv}_{1\times1}
\left(
\operatorname{DWConv}_{3\times3}
\left(F_i\right)
\right),
\end{equation}
where \(O_i\) denotes the objectness logit at the \(i\)-th scale. After the sigmoid function, the class-agnostic objectness probability is obtained as
\begin{equation}
P_{\mathrm{obj},i}
=
\sigma\left(O_i\right).
\end{equation}

During inference, OCDetect uses \(P_{\mathrm{obj},i}\) to calibrate the class confidence:
\begin{equation}
\widehat{P}_{\mathrm{cls},i}
=
P_{\mathrm{cls},i}
\cdot
\left(P_{\mathrm{obj},i}\right)^{\lambda},
\end{equation}
where \(\lambda\) denotes the confidence calibration strength coefficient. This calibration does not change the bounding-box regression result. Instead, it lowers the effective scores of false responses caused by background textures, desk-chair edges, and partial limbs before NMS, thereby reducing the number of invalid candidate boxes and improving post-processing stability.

During training, the objectness branch is supervised by labels generated from the task assignment results~\citep{zhang2020atss}. If a candidate position is assigned to a ground-truth object, its objectness label is set close to 1; otherwise, it is set to 0. The corresponding loss is defined as
\begin{equation}
\mathcal{L}_{\mathrm{obj}}
=
\operatorname{BCE}
\left(
P_{\mathrm{obj},i},
T_{\mathrm{obj},i}
\right),
\end{equation}
where \(T_{\mathrm{obj},i}\) denotes the objectness supervision label at the \(i\)-th scale. By introducing an independent objectness judgment, OCDetect decouples whether a real object exists from which category it belongs to, thereby alleviating the confusion between category responses and background-noise responses in highly dense occlusion regions.

\section{Experiments}\label{sec:experiments}

\subsection{Experimental Setup}\label{subsec:experimental_setup}

\subsubsection{Implementation Details}\label{subsubsec:implementation_details}
The proposed method is implemented based on the Ultralytics YOLO framework~\citep{jocher2023ultralytics} in the Visual Studio Code integrated development environment. The training program is executed on a computing platform running Ubuntu~22.04 Linux, equipped with two NVIDIA RTX~3090 GPUs, each with 24~GB memory. The software environment includes Python~3.11.14, PyTorch~2.2.2, CUDA~12.1, cuDNN~8.9.7, OpenCV~4.9.0, and Ultralytics~8.3.63. The main hyperparameters used for model training are listed in Table~4
\begin{table}[htbp]
\centering
\caption{Main hyperparameters used for model training.}
\label{tab:training_hyperparameters}
\small
\setlength{\tabcolsep}{8pt}
\begin{tabular}{lc}
\hline
Hyperparameter & Value \\
\hline
Learning rate & 0.01 \\
Decay strategy & Linear decay \\
Optimizer & Auto \\
Momentum & 0.937 \\
Weight decay & 0.0005 \\
Total epochs & 250 \\
Batch size & 32 \\
NMS threshold & 0.7 \\
Image size & 640 \\
Close mosaic epochs & 10 \\
\hline
\end{tabular}
\end{table}

\subsubsection{Datasets}\label{subsubsec:experimental_datasets}
SCB-D3-S~\citep{yang2023scbdataset3} is a public classroom behavior detection dataset. In this paper, it is used as an external validation dataset to evaluate the generalization ability of the proposed model in conventional classroom scenarios. In addition to the two main datasets, SBD~\citep{sun2021student}, STBD-08~\citep{zhao2023cbph}, and CrowdHuman~\citep{shao2018crowdhuman} are further introduced in the dataset benchmark difficulty validation in Section~5.2 for horizontal comparison. This comparison aims to analyze the differences among different datasets in terms of instance density, candidate-box quality, and detection difficulty.
\subsubsection{Evaluation Metrics}\label{subsubsec:evaluation_metrics}
This paper adopts \(\mathrm{mAP}_{50:95}\), \(\mathrm{mAP}_{50}\), \(\mathrm{AP}_{75}\)~\citep{lin2014coco}, Precision~(P), and Recall~(R) as the main detection metrics. Among them, \(\mathrm{mAP}_{50:95}\) is used as the primary evaluation metric, \(\mathrm{mAP}_{50}\) measures the basic detection ability, and \(\mathrm{AP}_{75}\) reflects the detection quality under a stricter localization threshold. Precision reflects false-positive suppression, whereas Recall reflects missed-detection coverage. To analyze candidate-box quality in highly congested scenarios, this paper further adopts Pre-NMS, \(\mathrm{FPPI}_{\mathrm{pre}}\), \(\mathrm{FPPI}\), \(\mathrm{FP}_{\mathrm{nei}}\), and \(\mathrm{FPPI}_{\mathrm{nei}}\). Pre-NMS~\citep{bodla2017softnms} denotes the average number of candidate boxes before non-maximum suppression. \(\mathrm{FPPI}_{\mathrm{pre}}\) denotes the average number of false positive candidate boxes per image before NMS. Unless otherwise specified, \(\mathrm{FPPI}\) denotes the average number of false positives per image after standard post-processing. \(\mathrm{FP}_{\mathrm{nei}}\) denotes the number of false positives caused by neighboring instances or occlusion boundaries. \(\mathrm{FPPI}_{\mathrm{nei}}\) denotes the average number of neighboring-instance or occlusion-boundary false positives per image after standard post-processing. In addition, \(\mathrm{AP}_{\mathrm{occ}}\), \(\mathrm{AP}_{75,\mathrm{occ}}\), \(\mathrm{AP}_{\mathrm{SF}}\), and \(\mathrm{R}_{\mathrm{SF}}\) are used to analyze the detection performance of occluded targets and distant small targets, while Params and FLOPs are used to measure model complexity.
\subsection{Benchmark Difficulty Verification of the Dataset}\label{subsec:benchmark_difficulty_verification}
To evaluate the benchmark difficulty of HCCB, this paper conducts a unified evaluation of YOLO and ODER-HSFNet on multiple classroom behavior and dense crowd datasets, as shown in Table~5. STBD-08 achieves the highest overall performance, suggesting that its target structures are relatively easier to model under this evaluation setting. In contrast, YOLOv13 only achieves 57.39\% \(\mathrm{mAP}_{50:95}\) and 76.70\% \(\mathrm{mAP}_{50}\) on HCCB, which is clearly lower than its performance on STBD-08. This indicates that high instance density, occlusion, and depth-induced scale discontinuities significantly increase the detection difficulty. After introducing ODER-HSFNet, \(\mathrm{mAP}_{50:95}\) on HCCB increases to 60.60\%, and Recall~(R) improves from 72.87\% to 78.91\%. These results show that the proposed method can alleviate part of the missed-detection problem, while HCCB still remains a highly challenging benchmark.

\begin{table}[width=.96\linewidth,cols=6,pos=htbp]
\centering
\caption{Detection performance comparison between YOLOv13 and ODER-HSFNet on different datasets.}
\label{tab:detection_performance_comparison}
\small
\setlength{\tabcolsep}{5pt}
\begin{tabular}{llccccc}
\hline
Dataset & Model & \(\mathrm{mAP}_{50:95}\) & \(\mathrm{mAP}_{50}\) & \(\mathrm{AP}_{75}\) & \(\mathrm{P}\) & \(\mathrm{R}\) \\
\hline
 & YOLOv13 & 48.94 & 72.14 & 55.90 & 74.78 & 68.69 \\
\rowcolor{gray!10}
\cellcolor{white}\multirow{-2}{*}{SBD} & ODER-HSFNet & 50.68 & 73.34 & 59.07 & 75.78 & 69.83 \\
\hline
 & YOLOv13 & 77.53 & 92.15 & 88.99 & 88.29 & 86.97 \\
\rowcolor{gray!10}
\cellcolor{white}\multirow{-2}{*}{STBD-08} & ODER-HSFNet & 78.38 & 92.33 & 89.06 & 88.18 & 87.55 \\
\hline
 & YOLOv13 & 51.39 & 70.07 & 58.77 & 68.54 & 64.98 \\
\rowcolor{gray!10}
\cellcolor{white}\multirow{-2}{*}{SCB-D3-S} & ODER-HSFNet & 55.64 & 73.13 & 63.56 & 70.57 & 68.12 \\
\hline
 & YOLOv13 & 53.85 & 83.96 & 57.82 & 85.70 & 73.62 \\
\rowcolor{gray!10}
\cellcolor{white}\multirow{-2}{*}{CrowdHuman} & ODER-HSFNet & 54.98 & 84.31 & 59.56 & 86.03 & 73.67 \\
\hline
 & YOLOv13 & 57.39 & 76.70 & 67.69 & 73.82 & 72.87 \\
\rowcolor{gray!10}
\cellcolor{white}\multirow{-2}{*}{HCCB (ours)} & ODER-HSFNet & 60.60 & 80.12 & 70.92 & 74.69 & 78.91 \\
\hline
\end{tabular}
\end{table}

Table~6 presents the comparison of Pre-NMS candidate-box quality. On HCCB, YOLOv13 generates 1520.55 Pre-NMS candidate boxes, 223.12 \(\mathrm{FPPI}\), and 24.23 \(\mathrm{FP}_{\mathrm{nei}}\), all of which are the highest among the compared datasets. This indicates that HCCB causes severe candidate-box congestion and neighboring-instance false positives. Even with ODER-HSFNet, the Pre-NMS and \(\mathrm{FPPI}\) values on HCCB still reach 1329.25 and 197.57, respectively, which are significantly higher than those on the other classroom behavior datasets. Meanwhile, \(\mathrm{FP}_{\mathrm{nei}}\) decreases from 24.23 to 18.71, showing that ODER-HSFNet can suppress part of the false positives caused by neighboring instances and occlusion boundaries. These results indicate that the difficulty of HCCB is reflected not only in the degradation of final detection accuracy, but also in the competition among a large number of low-quality candidate boxes at the early candidate generation stage.

\begin{table}[width=.94\linewidth,cols=5,pos=htbp]
\centering
\caption{Candidate-box quality and false-positive comparison on different datasets.}
\label{tab:candidate_box_quality}
\small
\setlength{\tabcolsep}{5pt}
\begin{tabular}{llccc}
\hline
Dataset & Model & Pre-NMS~\(\downarrow\) & \(\mathrm{FPPI}_{\mathrm{pre}}\,\downarrow\) & \(\mathrm{FP}_{\mathrm{nei}}\,\downarrow\) \\
\hline
 & YOLOv13 & 871.00 & 86.49 & 9.12 \\
\rowcolor{gray!10}
\cellcolor{white}\multirow{-2}{*}{SBD} & ODER-HSFNet & 604.93 & 50.45 & 5.51 \\
\hline
 & YOLOv13 & 583.14 & 47.40 & 4.19 \\
\rowcolor{gray!10}
\cellcolor{white}\multirow{-2}{*}{STBD-08} & ODER-HSFNet & 480.01 & 28.15 & 3.72 \\
\hline
 & YOLOv13 & 269.94 & 41.88 & 1.93 \\
\rowcolor{gray!10}
\cellcolor{white}\multirow{-2}{*}{SCB-D3-S} & ODER-HSFNet & 117.44 & 10.29 & 1.13 \\
\hline
 & YOLOv13 & 666.07 & 153.34 & 5.12 \\
\rowcolor{gray!10}
\cellcolor{white}\multirow{-2}{*}{CrowdHuman} & ODER-HSFNet & 398.54 & 69.74 & 1.47 \\
\hline
 & YOLOv13 & 1520.55 & 223.12 & 24.23 \\
\rowcolor{gray!10}
\cellcolor{white}\multirow{-2}{*}{HCCB (ours)} & ODER-HSFNet & 1329.25 & 197.57 & 18.71 \\
\hline
\end{tabular}
\vspace{2pt}
\begin{minipage}{\linewidth}
\footnotesize
Note: \(\mathrm{FPPI}_{\mathrm{pre}}\) is computed at the pre-NMS candidate-box level and measures the average number of false positive candidate boxes per image before non-maximum suppression.
\end{minipage}
\end{table}

Fig.~15 shows the normalized confusion matrices of YOLOv13 and ODER-HSFNet on HCCB. YOLOv13 exhibits clear confusion among fine-grained behavior categories, such as Reading, Writing, Bowing Head, and Using Phone, with the confusion among head-down-related categories being particularly prominent. After introducing ODER-HSFNet, the diagonal entries of most categories are enhanced. Specifically, the normalized correct classification responses of Reading, Writing, Heads Up, Bowing Head, and Using Phone increase to 0.71, 0.68, 0.87, 0.82, and 0.77, respectively. Meanwhile, the proportion of ground-truth targets predicted as Background is noticeably reduced. These results indicate that ODER-HSFNet can alleviate both category confusion and missed detections on HCCB.

\setcounter{figure}{14}
\begin{figure}
\centering
\includegraphics[width=.76\linewidth]{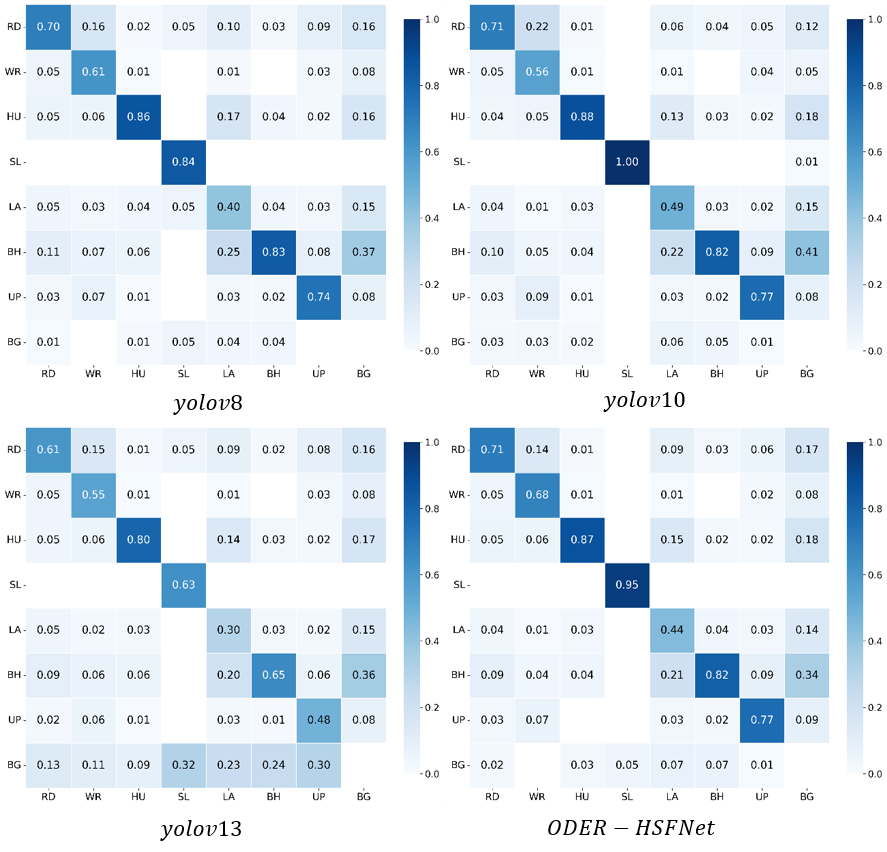}
\caption{Comparison of normalized confusion matrices for classroom action recognition. The categories include Reading (RD), Writing (WR), Heads Up (HU), Sleeping (SL), Looking Around (LA), Bowing Head (BH), Using Phone (UP), and Background (BG).}
\label{fig:confusion_matrices}
\end{figure}

\subsection{Comparison with Mainstream Methods}\label{subsec:mainstream_method_comparison}
To evaluate the overall performance of ODER-HSFNet, this paper compares it with mainstream YOLO-series detectors on HCCB and SCB-D3-S, as shown in Table~7. On HCCB, ODER-HSFNet achieves 60.60\% \(\mathrm{mAP}_{50:95}\) and 80.12\% \(\mathrm{mAP}_{50}\), obtaining the best results among all compared methods. Compared with the best-performing YOLO baseline, ODER-HSFNet improves \(\mathrm{mAP}_{50:95}\) and \(\mathrm{mAP}_{50}\) by 0.61 and 0.43 percentage points, respectively. On SCB-D3-S, ODER-HSFNet also achieves the highest performance, with \(\mathrm{mAP}_{50:95}\) and \(\mathrm{mAP}_{50}\) reaching 57.36\% and 74.65\%, respectively. These results correspond to improvements of 1.53 and 0.99 percentage points over the best-performing YOLO baseline. The results demonstrate that ODER-HSFNet achieves competitive detection performance in both highly congested classroom scenarios and conventional classroom scenarios.

\begin{table}[htbp]
\centering
\caption{Comparison with mainstream YOLO-series detectors on HCCB and SCB-D3-S.}
\label{tab:yolo_series_comparison}
\small
\setlength{\tabcolsep}{4pt}
\begin{tabular}{lcccccc}
\hline
\multirow{2}{*}{Model} & \multicolumn{2}{c}{HCCB} & \multicolumn{2}{c}{SCB-D3-S} & \multirow{2}{*}{Params~(M)} & \multirow{2}{*}{FLOPs~(G)} \\
\cline{2-5}
& \(\mathrm{mAP}_{50:95}\) & \(\mathrm{mAP}_{50}\) & \(\mathrm{mAP}_{50:95}\) & \(\mathrm{mAP}_{50}\) &  &  \\
\hline
YOLOv6s~\citep{li2022yolov6} & 58.32 & 78.696 & 55.74 & 73.66 & 16.31 & 43.90 \\
YOLOv8s~\citep{terven2023yolo} & 59.77 & 79.27 & 54.68 & 72.64 & 11.14 & 28.66 \\
YOLOv9s~\citep{wang2024yolov9} & 59.99 & 78.85 & 55.04 & 71.37 & 7.29 & 27.39 \\
YOLOv10s~\citep{wang2024yolov10} & 59.73 & 78.69 & 53.94 & 71.66 & 8.07 & 24.80 \\
YOLOv11s~\citep{khanam2024yolov11} & 59.95 & 79.69 & 55.26 & 73.40 & 9.43 & 21.56 \\
YOLOv12s~\citep{tian2025yolov12} & 57.63 & 76.53 & 55.83 & 73.20 & 9.10 & 19.59 \\
YOLOv13s~\citep{lei2025yolov13} & 57.39 & 76.70 & 51.39 & 70.07 & 9.03 & 21.00 \\
YOLOv26s~\citep{jocher2026yolo26} & 59.15 & 78.64 & 54.18 & 72.34 & 9.39 & 21.26 \\
\rowcolor{gray!10}
ODER-HSFNet &  \textbf{60.60} &  \textbf{80.12} &  \textbf{57.36} &  \textbf{74.65} & 12.74 & 24.87 \\
\hline
\end{tabular}
\end{table}

Fig.~16 presents the detection visualization results of different models. Each row corresponds to one classroom sample, and the columns show the original image, YOLOv10s, YOLOv13s, and ODER-HSFNet, respectively. The upper-left corner of each subfigure indicates the original image or the model name, while the upper-right corner provides instance-level statistics. Here, Total/Box denotes the number of ground-truth annotations or boxes, Pred denotes the number of predicted boxes, Box-Cls denotes the number of detections with both correct localization and correct classification, and FP denotes the number of false positives.
\setcounter{figure}{15}
\renewcommand{\theHfigure}{qualitative.\arabic{figure}}
\begin{figure}
\centering
\includegraphics[width=.98\linewidth]{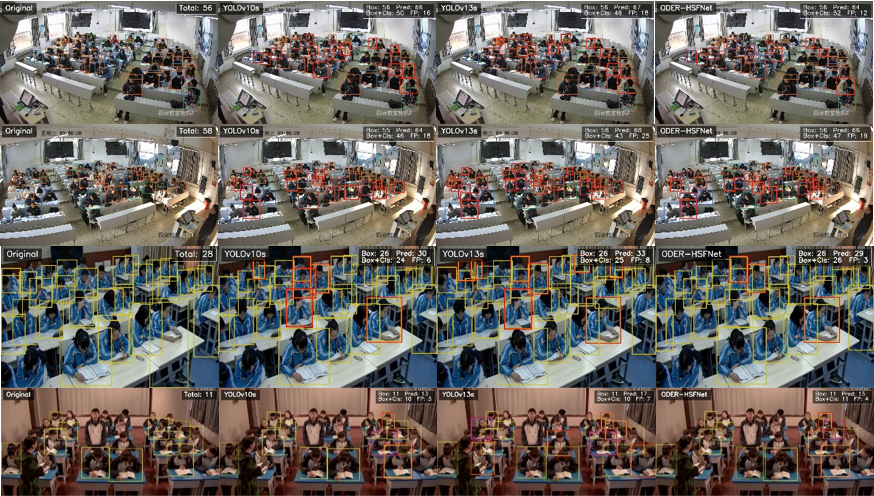}
\caption{Qualitative comparison of classroom behavior detection on the HCCB and SCB-D3-S datasets. The upper-left label indicates the original image or detector, and the upper-right label summarizes instance statistics: Total/Box denotes ground-truth annotations, Pred denotes predicted boxes, Box-Cls denotes correctly localized and classified detections, and FP denotes false positives.}
\label{fig:qualitative_detection_comparison}
\end{figure}
\renewcommand{\theHfigure}{\arabic{figure}}
In the first HCCB sample, the original image contains 56 annotated instances. YOLOv10s obtains Box 56, Pred 66, Box-Cls 50, and FP 16, while YOLOv13s obtains Box 56, Pred 67, Box-Cls 49, and FP 18. In contrast, ODER-HSFNet obtains Box 56, Pred 64, Box-Cls 52, and FP 12. This indicates that ODER-HSFNet achieves more correct detections and produces fewer false positives in this dense classroom scenario. In the third sample, YOLOv10s obtains Box 26, Pred 30, Box-Cls 24, and FP 6, while YOLOv13s obtains Box 26, Pred 33, Box-Cls 25, and FP 8. ODER-HSFNet obtains Box 26, Pred 29, Box-Cls 26, and FP 3, showing more stable localization and classification results in close-range occlusion and local behavior regions.

Overall, compared with YOLOv10s and YOLOv13s, ODER-HSFNet detects more valid student instances in densely occluded regions and reduces missed detections of rear-row small targets and neighboring-instance false positives. In summary, ODER-HSFNet achieves the best overall performance on both evaluated datasets among the compared methods, demonstrating its effectiveness for highly dense and occluded classroom behavior detection.
\setcounter{subsection}{3}
\subsection{Model Ablation Experiments}\label{subsec:model_ablation_experiments}

\subsubsection{ODER Module Component Ablation Results}\label{subsubsec:oder_ablation}
To evaluate the design effectiveness of ODER, this paper conducts ablation analysis from five aspects: feature-layer insertion position, internal components, sampling constraints, occluded-object detection, and distant small-object detection. All experiments adopt the same training settings, and only the configuration under verification is changed.

First, Table~8 presents the effect of inserting ODER into different feature layers. Among the single-layer configurations, \(X^2\) achieves 57.92\% \(\mathrm{mAP}_{50:95}\) and 77.70\% \(\mathrm{mAP}_{50}\) on HCCB, outperforming \(X^1\) and \(X^3\). This indicates that middle-level features are more responsive to occlusion boundaries and local behavior cues. Among the two-layer configurations, the \(X^1,X^2\) configuration achieves 58.31\% \(\mathrm{mAP}_{50:95}\) and 77.97\% \(\mathrm{mAP}_{50}\) on HCCB, and 52.90\% \(\mathrm{mAP}_{50:95}\) and 71.22\% \(\mathrm{mAP}_{50}\) on SCB-D3-S, obtaining the best overall performance. Compared with the \({X^1,X^3}\) and \({X^2,X^3}\) configurations, \({X^1,X^2}\) better balances shallow details and middle-level semantics. After further adding \(X^3\), \(\mathrm{mAP}_{50}\) does not continue to improve, while the number of parameters and FLOPs increase. Therefore, this paper adopts \({X^1,X^2}\) as the final insertion position of ODER.

\begin{table}[width=.98\linewidth,cols=7,pos=htbp]
\centering
\caption{Ablation study on the insertion position of ODER.}
\label{tab:oder_insertion_position}
\small
\setlength{\tabcolsep}{4pt}
\begin{tabular}{lcccccc}
\hline
\multirow{2}{*}{Design} & \multicolumn{2}{c}{HCCB} & \multicolumn{2}{c}{SCB-D3-S} & \multirow{2}{*}{Params~(M)} & \multirow{2}{*}{FLOPs~(G)} \\
\cline{2-5}
& \(\mathrm{mAP}_{50:95}\) & \(\mathrm{mAP}_{50}\) & \(\mathrm{mAP}_{50:95}\) & \(\mathrm{mAP}_{50}\) &  &  \\
\hline
\(X^1\) & 57.90 & 76.88 & 51.14 & 69.72 & 9.19 & 23.12 \\
\(X^2\) & 57.92 & 77.70 & 51.89 & 70.62 & 9.19 & 21.52 \\
\(X^3\) & 57.82 & 76.64 & 51.33 & 70.11 & 9.67 & 21.50 \\
\rowcolor{gray!10}
\({X^1,X^2}\) (ours) &  \textbf{58.31} &  \textbf{77.97} &  \textbf{52.90} &  \textbf{71.22} & 9.36 & 23.65 \\
\({X^1,X^3}\) & 57.92 & 77.18 & 52.75 & 71.10 & 9.84 & 23.63 \\
\({X^2,X^3}\) & 57.01 & 77.19 & 51.53 & 69.75 & 9.84 & 22.04 \\
\({X^1,X^2,X^3}\) &  \textbf{58.31} & 77.69 & 52.56 & 70.88 & 10.00 & 24.16 \\
\hline
\end{tabular}
\end{table}
Table~9 presents the ablation results of the internal components of ODER. When only DEAO is adopted, the model achieves 57.60\% \(\mathrm{mAP}_{50:95}\) and 77.70\% \(\mathrm{mAP}_{50}\) on HCCB, indicating that deformable edge aggregation~\citep{dai2017deformable} can already enhance local occlusion evidence. After introducing HDSG, \(\mathrm{mAP}_{50:95}\) increases to 58.12\%, demonstrating that topology-aware sampling routing helps stabilize the sampling process in highly dense regions. When HDSG is replaced with a standard \(1\times1\) mapping, the performance is lower than that of the complete structure, suggesting that simple channel mapping cannot effectively substitute for structured sampling routing. The complete ODER achieves 58.31\% and 52.90\% \(\mathrm{mAP}_{50:95}\) on HCCB and SCB-D3-S, respectively, obtaining the best overall performance.

\begin{table}[width=.94\linewidth,cols=5,pos=htbp]
\centering
\caption{Ablation study on the internal components of ODER.}
\label{tab:oder_component_ablation}
\small
\setlength{\tabcolsep}{5pt}
\begin{tabular}{lccccccc}
\hline
Dataset & DEAO & HDSG & Scale & \(\mathrm{mAP}_{50:95}\) & \(\mathrm{mAP}_{50}\) & \(\mathrm{P}\) & \(\mathrm{R}\) \\
\hline
HCCB & \(\checkmark\) & \(\times\) & \(\times\) & 57.60 & 77.70 & 73.25 & 74.17 \\
HCCB & \(\checkmark\) & \(\checkmark\) & \(\times\) & 58.12 & 77.55 & 71.69 & 74.79 \\
HCCB & \(\checkmark\) & \(\triangle\) & \(\checkmark\) & 57.71 & 77.34 & 69.39 & 77.69 \\
\rowcolor{gray!10}
HCCB & \(\checkmark\) & \(\checkmark\) & \(\checkmark\) &  \textbf{58.31} &  \textbf{77.97} &  \textbf{73.33} &  \textbf{75.03} \\
\hline
SCB-D3-S & \(\checkmark\) & \(\times\) & \(\times\) & 51.99 & 70.64 & 66.56 & 66.27 \\
SCB-D3-S & \(\checkmark\) & \(\checkmark\) & \(\times\) & 51.54 & 69.90 & 66.19 & 66.80 \\
SCB-D3-S & \(\checkmark\) & \(\triangle\) & \(\checkmark\) & 52.24 & 70.28 &  \textbf{67.66} & 65.95 \\
\rowcolor{gray!10}
SCB-D3-S & \(\checkmark\) & \(\checkmark\) & \(\checkmark\) &  \textbf{52.90} &  \textbf{71.22} & 65.84 &  \textbf{67.69} \\
\hline
\end{tabular}

\vspace{1mm}
\begin{minipage}{\linewidth}
\footnotesize
Note: \(\triangle\) denotes that HDSG is replaced with a standard \(1\times1\) mapping.
\end{minipage}
\end{table}

Table~10 presents a sensitivity analysis of the number of sampling hypotheses \(K\) and the \(\tanh\)-based bounded offset constraint in ODER. Without the \(\tanh\) constraint, \(K=4\) achieves a relatively high \(\mathrm{mAP}_{50:95}\) on HCCB, but its performance on SCB-D3-S is weaker, indicating limited cross-dataset stability when unbounded offsets are used. As \(K\) increases to 8 or 12, the performance of unconstrained sampling further degrades, suggesting that excessive unconstrained sampling may introduce interference from neighboring instances or background regions. After applying the \(\tanh\)-based constraint, the model exhibits more stable performance. In particular, \(K=8\) achieves \(\mathrm{mAP}_{50:95}\) values of 58.31\% and 52.90\% on HCCB and SCB-D3-S, respectively, yielding the best overall trade-off across the two datasets. Therefore, \(K=8\) with the \(\tanh\)-based bounded offset constraint is adopted in the final configuration.

Table~10 reports the sensitivity analysis of the number of sampling hypotheses \(K\) and the \(\tanh\)-based bounded offset constraint in ODER, where \(\tanh\) indicates whether bounded constraints are applied to the sampling offsets.

\begin{table}[width=.96\linewidth,cols=6,pos=htbp]
\centering
\caption{Ablation study on the sampling number \(K\) and \(\tanh\) constraint in ODER.}
\label{tab:oder_sampling_constraint}
\small
\setlength{\tabcolsep}{6pt}
\begin{tabular}{cccccc}
\hline
\multirow{2}{*}{\(K\)} & \multirow{2}{*}{\(\tanh\)} & \multicolumn{2}{c}{HCCB} & \multicolumn{2}{c}{SCB-D3-S} \\
\cline{3-6}
&  & \(\mathrm{mAP}_{50:95}\) & \(\mathrm{mAP}_{50}\) & \(\mathrm{mAP}_{50:95}\) & \(\mathrm{mAP}_{50}\) \\
\hline
1  & \(\times\)      & 57.69 & 77.14 & 51.92 & 70.05 \\
4  & \(\times\)      & \textbf{58.50} & 77.07 & 51.63 & 69.85 \\
8  & \(\times\)      & 57.86 & 77.20 & 50.77 & 68.91 \\
12 & \(\times\)      & 57.10 & 76.30 & 50.83 & 69.40 \\
\hline
1  & \(\checkmark\) & 58.45 & 77.28 & 52.27 & 70.86 \\
4  & \(\checkmark\) & 58.07 & 77.13 & 51.59 & 69.76 \\
\rowcolor{gray!10}
8  & \(\checkmark\) & 58.31 & \textbf{77.97} & \textbf{52.90} & \textbf{71.22} \\
12 & \(\checkmark\) & 57.66 & 77.31 & 52.33 & 71.04 \\
\hline
\end{tabular}
\end{table}

Figure~17 shows the boundary-path visualization of ODER in partially occluded scenarios. When interference is introduced by adjacent students, desk and chair edges, or local body parts, ODER performs controlled sampling along the visible target boundaries, preventing the sampling path from excessively extending into neighboring instance regions. These results indicate that ODER strengthens effective boundary evidence for occluded targets and reduces feature contamination in high-density regions.

\setcounter{figure}{16}
\begin{figure}
\centering
\includegraphics[width=.98\linewidth]{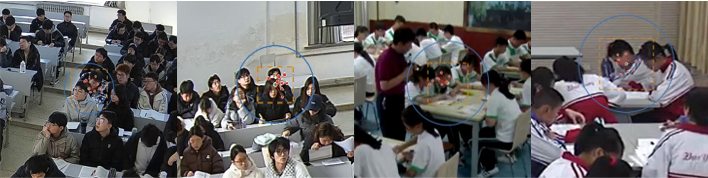}
\caption{Visualization of the ODER boundary-guided route in partially occluded scenes. The blue circles highlight local regions with occlusion interference, the orange dashed boxes indicate occluded targets, and the yellow routes represent the boundary-guided responses generated by ODER.}
\label{fig:oder_boundary_guided_route}
\end{figure}

Table~11 reports the results of occluded-target detection and neighboring false-positive suppression. On HCCB, ODER improves \(\mathrm{AP}_{\mathrm{occ}}\) from 46.179\% to 47.572\% and \(\mathrm{AP}_{75,\mathrm{occ}}\) from 35.100\% to 36.374\%, while reducing \(\mathrm{FP}_{\mathrm{nei}}\) and \(\mathrm{FPPI}_{\mathrm{nei}}\). On SCB-D3-S, ODER also improves \(\mathrm{AP}_{\mathrm{occ}}\) and \(\mathrm{AP}_{75,\mathrm{occ}}\), with fewer neighboring false positives. These results indicate that ODER enhances the detection of occluded targets and suppresses false-positive responses caused by adjacent students and occlusion boundaries.

\begin{table}[width=.94\linewidth,cols=5,pos=htbp]
\centering
\caption{Ablation study of ODER on occlusion-aware detection and neighboring false positives.}
\label{tab:oder_occlusion_fp}
\small
\setlength{\tabcolsep}{6pt}
\begin{tabular}{llcccc}
\hline
Dataset & Model & \(\mathrm{AP}_{\mathrm{occ}}\,\uparrow\) & \(\mathrm{AP}_{75,\mathrm{occ}}\,\uparrow\) & \(\mathrm{FP}_{\mathrm{nei}}\,\downarrow\) & \(\mathrm{FPPI}_{\mathrm{nei}}\,\downarrow\) \\
\hline
 & YOLOv13 & 46.179 & 35.100 & 393.960 & 0.983 \\
\rowcolor{gray!10}
\cellcolor{white}\multirow{-2}{*}{HCCB} & ODER & 47.572 & 36.374 & 390.902 & 0.972 \\
\hline
 & YOLOv13 & 36.642 & 25.699 & 362.564 & 0.978 \\
\rowcolor{gray!10}
\cellcolor{white}\multirow{-2}{*}{SCB-D3-S} & ODER & 38.937 & 27.280 & 356.374 & 0.975 \\
\hline
\end{tabular}
\vspace{2pt}
\begin{minipage}{\linewidth}
\footnotesize
Note: \(\mathrm{FPPI}_{\mathrm{nei}}\) is computed after standard post-processing and is used to evaluate residual neighboring false positives around adjacent or occluded instances.
\end{minipage}
\end{table}

Table~12 compares the performance of ODER on distant small targets. On HCCB, ODER improves \(\mathrm{AP}_{\mathrm{SF}}\) from 41.727\% to 43.480\% and \(\mathrm{R}_{\mathrm{SF}}\) from 79.780\% to 80.678\%. On SCB-D3-S, \(\mathrm{AP}_{\mathrm{SF}}\) and \(\mathrm{R}_{\mathrm{SF}}\) are also improved to 34.785\% and 71.309\%, respectively. These results suggest that ODER alleviates the loss of visible evidence for distant small targets caused by low resolution, occlusion, and ambiguous boundaries.

\begin{table}[width=.98\linewidth,cols=7,pos=htbp]
\centering
\caption{Performance comparison of ODER on distant small targets.}
\label{tab:oder_small_far}
\small
\setlength{\tabcolsep}{8pt}
\begin{tabular}{cccc}
\hline
Dataset & Model & \(\mathrm{AP}_{\mathrm{SF}}\,\uparrow\) & \(\mathrm{R}_{\mathrm{SF}}\,\uparrow\) \\
\hline
\multirow{2}{*}{HCCB}
& YOLOv13 & 41.727 & 79.780 \\
& ODER & 43.480 & 80.678 \\
\hline
\multirow{2}{*}{SCB-D3-S}
& YOLOv13 & 33.567 & 69.761 \\
& ODER & 34.785 & 71.309 \\
\hline
\end{tabular}
\end{table}

Overall, the optimal configuration of ODER introduces the complete DEAO-HDSG-Scale structure into the \(X^1\) and \(X^2\) feature layers, and adopts \(K=8\) with the \(\tanh\)-based bounded offset constraint. The experimental results demonstrate the effectiveness of ODER in occlusion evidence compensation, neighboring false-positive suppression, and distant small-target detection in highly crowded classroom scenarios.

\subsubsection{HSSF Module Component Ablation Results}\label{subsubsec:hssf_ablation}
To evaluate the effectiveness of HSSF, ablation experiments are conducted from three aspects: component contribution, feature reconstruction strategy, and cross-scale response quality. All experiments use ODER as the baseline and follow the same training settings.

Table~13 presents the ablation results of different HSSF components. Compared with the ODER baseline, Base Flow improves \(\mathrm{mAP}_{50:95}\) on HCCB from 58.31\% to 58.94\%, indicating that basic cross-scale reconstruction~\citep{lin2017fpn,tan2020efficientdet} provides a measurable performance gain. After introducing the VSS Proxy~\citep{gu2023mamba,liu2024vmamba}, \(\mathrm{mAP}_{50:95}\) increases to 59.67\% on HCCB and 55.25\% on SCB-D3-S, suggesting that state-space contextual modeling facilitates cross-region information interaction in dense scenes. When Hypergraph is introduced alone, the performance on HCCB decreases, indicating that high-order relation modeling may introduce relational noise without stable local and global feature support. The complete HSSF achieves 60.34\% \(\mathrm{mAP}_{50:95}\) and 79.96\% \(\mathrm{mAP}_{50}\) on HCCB, and 55.61\% \(\mathrm{mAP}_{50:95}\) and 73.58\% \(\mathrm{mAP}_{50}\) on SCB-D3-S, obtaining the best overall performance across the two datasets.

\begin{table}[width=.98\linewidth,cols=8,pos=htbp]
\centering
\caption{Ablation study on different design variants in HSSF.}
\label{tab:hssf_component_ablation}
\small
\setlength{\tabcolsep}{6pt}
\begin{tabular}{cccccccc}
\hline
\multirow{2}{*}{Step} & \multirow{2}{*}{Design} & \multicolumn{2}{c}{HCCB} & \multicolumn{2}{c}{SCB-D3-S} & \multirow{2}{*}{Params~(M)} & \multirow{2}{*}{FLOPs~(G)} \\
\cline{3-6}
& & \(\mathrm{mAP}_{50:95}\) & \(\mathrm{mAP}_{50}\) & \(\mathrm{mAP}_{50:95}\) & \(\mathrm{mAP}_{50}\) & & \\
\hline
0 & ODER & 58.31 & 77.97 & 52.90 & 71.22 & 9.36 & 23.65 \\
1 & Base flow & 58.94 & 77.90 & 53.27 & 71.95 & 8.79 & 22.90 \\
2 & VSS proxy & 59.67 & 79.34 & 55.25 & 73.43 & 12.18 & 24.22 \\
3 & Hypergraph & 57.99 & 77.12 & 53.72 & 72.74 & 9.32 & 23.53 \\
4 & Local flow & 58.23 & 77.58 & 53.27 & 71.89 & 8.83 & 23.03 \\
\rowcolor{gray!10}
5 & HSSF (Ours) & \textbf{60.34} & \textbf{79.96} & \textbf{55.61} & \textbf{73.58} & 12.74 & 24.97 \\ 
\hline
\end{tabular}
\end{table}

Table~14 compares different feature reconstruction strategies. Sequential, Residual, Dynamic Residual, and Fission Routing achieve moderate improvements, but their performance is less stable across datasets. In contrast, the proposed HSSF strategy achieves \(\mathrm{mAP}_{50:95}\) values of 60.34\% and 55.61\% on HCCB and SCB-D3-S, respectively, both representing the best results among the compared strategies. These results indicate that HSSF provides a more effective balance between detail preservation and cross-scale semantic alignment.

\begin{table}[width=.96\linewidth,cols=6,pos=htbp]
\centering
\caption{Ablation study on different fusion designs in HSSF.}
\label{tab:hssf_fusion_designs}
\small
\setlength{\tabcolsep}{6pt}
\begin{tabular}{cccccc}
\hline
\multirow{2}{*}{Step} & \multirow{2}{*}{Design} & \multicolumn{2}{c}{HCCB} & \multicolumn{2}{c}{SCB-D3-S} \\
\cline{3-6}
& & \(\mathrm{mAP}_{50:95}\) & \(\mathrm{mAP}_{50}\) & \(\mathrm{mAP}_{50:95}\) & \(\mathrm{mAP}_{50}\) \\
\hline
0 & Sequential & 60.15 & 79.32 & 55.25 & \textbf{73.68} \\
1 & Residual & 60.05 & 79.41 & 53.80 & 71.88 \\
2 & Dynamic residual & 60.02 & 79.40 & 55.11 & 73.19 \\
3 & Fission routing & 59.80 & 79.23 & 54.76 & 73.01 \\
\rowcolor{gray!10}
4 & HSSF (Ours) & \textbf{60.34} & \textbf{79.96} & \textbf{55.61} & 73.58 \\
\hline
\end{tabular}
\end{table}

Figure~18 presents the cross-scale response heatmaps of HSSF with adaptive hyperedges. HSSF produces more concentrated responses on student body regions across different scales, while suppressing activations in non-target regions such as desks, chairs, and background edges. This observation suggests that adaptive hyperedges~\citep{feng2019hypergraph} facilitate high-order cross-region association modeling, thereby improving the stability of target responses in highly crowded and occluded scenarios.

\begin{figure}
\centering
\includegraphics[width=.98\linewidth]{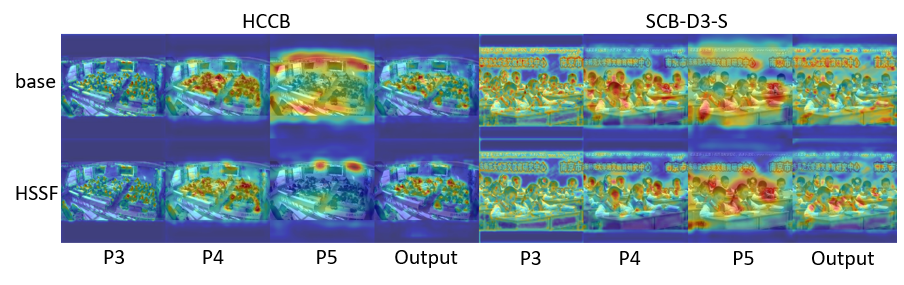}
\caption{Cross-scale response heatmap comparison between Adaptive Hyperedges and HSSF. Warmer colors indicate stronger feature responses. HSSF enhances responses on student instances and suppresses irrelevant background activations.}
\label{fig:hssf_response_heatmap}
\end{figure}

Table~15 reports the changes in the proportion of high-response pixels located within ground-truth (GT) target boxes. After introducing HSSF, the output features show relative increases of 12.9\% and 49.8\% on HCCB and SCB-D3-S, respectively. This indicates that HSSF improves the spatial consistency between feature responses and ground-truth student targets. In particular, the \(X^3\) layer achieves improvements of 8.4\% and 22.0\% on the two datasets, respectively, suggesting that HSSF provides stronger target alignment for deep semantic responses.

\begin{table}[width=.92\linewidth,cols=4,pos=htbp]
\centering
\caption{Changes in the proportion of high-response pixels located inside GT boxes.}
\label{tab:feature_stage_improvement}
\small
\renewcommand{\arraystretch}{1.15}
\setlength{\tabcolsep}{8pt}
\begin{tabular}{ccccc}
\hline
Dataset & \(X^1\) & \(X^2\) & \(X^3\) & Output \\
\hline
HCCB & +1.9\% & -3.1\% & +8.4\% & +12.9\% \\
SCB-D3-S & +7.4\% & +1.9\% & +22.0\% & +49.8\% \\
\hline
\end{tabular}
\end{table}

Overall, HSSF mitigates cross-scale feature misalignment and dense response competition in highly crowded classroom scenarios through the coordinated effects of local structure enhancement, state-space contextual modeling, and adaptive hypergraph relation aggregation. It also improves the response consistency of target regions.

\subsubsection{OCDetect Module Component Ablation Results}\label{subsubsec:ocdetect_ablation}

To evaluate the candidate-box calibration capability of OCDetect, the original detection head is replaced with OCDetect in YOLO and ODER+HSSF, respectively. The number of Pre-NMS candidate boxes, FPPI, and detection accuracy are compared, as reported in Table~16.

On HCCB, introducing OCDetect into YOLO reduces the number of Pre-NMS candidate boxes from 1521.59 to 1403.95 and \(\mathrm{FPPI}\) from 223.40 to 205.45, while increasing \(\mathrm{mAP}_{50:95}\) to 57.84\%. For ODER+HSSF, OCDetect further reduces Pre-NMS from 1591.76 to 1329.76 and \(\mathrm{FPPI}\) from 217.68 to 197.55, while improving \(\mathrm{mAP}_{50:95}\) to 60.60\%. These results indicate that OCDetect reduces low-quality candidate boxes in highly crowded classroom scenarios.

On SCB-D3-S, OCDetect also provides consistent improvements. For YOLO, Pre-NMS decreases from 270.28 to 119.09, \(\mathrm{FPPI}\) decreases from 41.91 to 11.04, and \(\mathrm{mAP}_{50:95}\) increases to 52.79\%. For ODER+HSSF, Pre-NMS decreases from 227.77 to 117.98, \(\mathrm{FPPI}\) decreases from 35.01 to 10.28, and \(\mathrm{mAP}_{50:95}\) increases to 57.36\%. These results suggest that OCDetect improves candidate-box ranking quality across different classroom scenarios.

Table~16 reports the ablation results of OCDetect for candidate-box calibration. Pre-NMS denotes the average number of candidate boxes before non-maximum suppression, and \(\mathrm{FPPI}\) denotes the average number of false positives per image, where lower values are better.

\begin{table}[width=.92\linewidth,cols=4,pos=htbp]
\centering
\caption{Ablation study of OCDetect under different model settings.}
\label{tab:ocdetect_ablation}
\small
\renewcommand{\arraystretch}{1.15}
\setlength{\tabcolsep}{8pt}
\begin{tabular}{cclcccc}
\hline
Dataset & Model & Head & Pre-NMS~\(\downarrow\) & \(\mathrm{FPPI}_{\mathrm{pre}}\,\downarrow\) & \(\mathrm{mAP}_{50:95}\) & \(\mathrm{mAP}_{50}\) \\
\hline
 &  & Detect   & 1521.59 & 223.40 & 57.39 & 76.70 \\
 & \cellcolor{white}\multirow{-2}{*}{YOLOv13} & OCDetect & 1403.95 & 205.45 & 57.84 & 77.27 \\
 &  & Detect   & 1591.76 & 217.68 & 60.34 & 79.96 \\
\rowcolor{gray!10}
\cellcolor{white}\multirow{-4}{*}{HCCB} & \cellcolor{white}\multirow{-2}{*}{ODER + HSSF} & OCDetect & \textbf{1329.76} & \textbf{197.55} & \textbf{60.60} & \textbf{80.12} \\
\hline
 &  & Detect   & 270.28 & 41.91 & 51.39 & 70.07 \\
 & \cellcolor{white}\multirow{-2}{*}{YOLOv13} & OCDetect & 119.09 & 11.04 & 52.79 & 70.45 \\
 &  & Detect   & 227.77 & 35.01 & 55.61 & 73.58 \\
\rowcolor{gray!10}
\cellcolor{white}\multirow{-4}{*}{SCB-D3-S} & \cellcolor{white}\multirow{-2}{*}{ODER + HSSF} & OCDetect & \textbf{117.98} & \textbf{10.28} & \textbf{57.36} & \textbf{74.65} \\
\hline
\end{tabular}
\vspace{2pt}
\begin{minipage}{\linewidth}
\footnotesize
Note: \(\mathrm{FPPI}_{\mathrm{pre}}\) is computed at the pre-NMS candidate-box level and measures the average number of false positive candidate boxes per image before non-maximum suppression.
\end{minipage}
\end{table}

Overall, OCDetect effectively reduces low-quality candidate boxes and consistently improves detection accuracy on HCCB and SCB-D3-S.

\section{Conclusion}\label{sec:conclusion}

This study focuses on student behavior detection in real-world large-class classroom scenarios. To address the limited coverage of dense instances, severe occlusion, and depth-wise scale conflicts in existing classroom behavior datasets, we construct HCCB, a highly crowded classroom behavior detection dataset. HCCB contains 796 high-resolution classroom images and 50,229 student behavior instances covering seven typical classroom behaviors. With higher instance density, stronger object-level occlusion, and more pronounced front-to-back scale discontinuity, HCCB provides a challenging benchmark for highly crowded classroom behavior detection.

Methodologically, we propose ODER-HSFNet based on the YOLO framework. ODER compensates for visible evidence around occluded boundaries and distant degraded regions, HSSF models local structures, state-space contexts, and high-order hypergraph relations, and OCDetect calibrates candidate-box quality at the detection head. These modules improve the adaptability of the model to highly crowded classroom scenarios from three complementary aspects: local evidence compensation, cross-scale relation fusion, and output confidence calibration.

Experimental results show that ODER-HSFNet outperforms mainstream YOLO-series methods on both HCCB and SCB-D3-S. Further candidate-box quality analysis, occluded-target evaluation, distant small-target evaluation, and module ablation experiments indicate that the proposed modules effectively alleviate neighboring false positives, missed detections of distant small targets in rear rows, and cross-scale response misalignment in densely occluded scenarios.

Future work will proceed in two directions. First, the scale and acquisition diversity of HCCB will be expanded to cover more classroom layouts, illumination conditions, and classroom activity types. Second, lightweight high-order relation modeling and video-based temporal behavior analysis will be further explored to improve the efficiency and continuous behavioral-state perception capability of intelligent classroom systems in practical deployment.

\printcredits

\bibliographystyle{unsrtnat}
\bibliography{cas-refs}



\end{document}